\documentstyle[twocolumn,named,a4wide,psfig]{article}

\newcommand{\epsfig}[3]{\centerline{\hbox{\psfig{figure=#1.eps,height=#2mm,width=#3mm}}}}

\begin{document}
\bibliographystyle{named}

{\onecolumn
\begin{center}
\large {\bf The Theoretical Status of Ontologies in Natural
Language Processing}\\[0.2in]
John A. Bateman\footnote{Most of the background for this paper is drawn from experiences with the
development of the Penman {\em  Upper Model}: an ontology  for supporting
natural language generation. The Upper Model has been under  development
since 1985, and many have and continue to contribute to it. The ideas  I
report on here would not have been possible without that  development.
Those responsible  for  the  present  form  of  the Upper Model include:
William Mann,  Christian  Matthiessen,  Robert  Kasper, Richard Whitney,
Johanna  Moore,  Eduard  Hovy,  Yigal  Arens,  and  myself.}\\
Projekt {\sc komet} and Penman Project\\  GMD/IPSI and USC/ISI \\
\normalsize \em e-mail: \tt bateman@gmd.de\\[0.2in]
March 5, 1992
\end{center}

{\small {\bf Note:}  \em This paper appears 
in the Proceedings of the workshop on `Text Representation and
Domain Modelling -- Ideas from Linguistics and AI', held at the Technical
University Berlin, October 9th - 11th, 1991. KIT Report 97, edited by Susanne
Preu\ss\ and Birte Schmitz.}}

Cmp-lg Paper No: cmp-lg/9704010

\twocolumn

\section{Introduction}

The development of natural  language processing (henceforth,  {\sc nlp})
systems has reached the stage  where concentrated efforts are  necessary
in the area  of representing  more `abstract',  more `knowledge'-related
bodies of information.   It has  been accepted that without  substantial
bodies  of  background  information  concerning  commonsense,   everyday
knowledge about the world or detailed information concerning  particular
domains of application,  it will  not be  possible to  construct systems
that can support the use of natural language. Systems need to  represent
concrete details  of  the  `worlds'  that  their  texts  describe:   for
example, the resolution of anaphors, the induction of text coherence  by
recognizing regularities present in the world  and not in the text,  the
recognition of  plans  by  knowing  what  kinds  of plans make sense for
speakers and  hearers  in  real  situations,  etc.    all  require world
modelling to various depths.

This need creates two interrelated problem areas.  The first problem  is
how knowledge of the world ---  be it general, commonsense knowledge  or
specialized knowledge concerning  some particular  domain ---  is to  be
represented. The second problem is  how such organizations of  knowledge
are to be related  to linguistic system  levels of organization  such as
grammar and lexis.         For  both problem areas  the concept of  {\bf
ontologies} for {\sc nlp} has been  suggested to be of potential  value.
Very generally,  an  ontology  offers  a  `conceptual' framework for the
representation of  information  ---  a  framework  that  is sufficiently
general, but also  sufficiently detailed,  to provide  a rich supportive
scaffolding for the construction of models  of the world. The design  of
such ontologies constitutes  an area  of concern  that is  coming to  be
known         as          {\em          ontological         engineering}
(e.g.,~\cite{NirenburgRaskin87,LenatGuha88,Simmons91}. As  we  shall see
below, most systems  that deal  currently with  {\sc nlp}  already adopt
some kind of  ontology for  their more  abstract levels  of information.
However, theoretical  principles  for  the  design  and  development  of
ontologies meeting the goals of generality and detail remain weak.  This
is due not only to a lack of theoretical accounts at these more rarified
abstract levels of information, but also to the co-existence of a  range
of,  sometimes   poorly   differentiated,   functions   such  bodies  of
information are expected to fulfill.

The following list gives an idea of the range of functions adopted in
{\sc nlp}. Ontologies are often expected to fulfill at least one (and
often more) of:
\begin{itemize}
\item organizing `world knowledge',
\item organizing the world itself,
\item organizing `meaning' or `semantics' of natural language
expressions,
\item providing an interface between system external components, domain
models, etc. and {\sc nlp} linguistic components,
\item ensuring expressability of input expressions,
\item offering an interlingua for machine translation,
\item supporting the construction of `conceptual dictionaries'.
\end{itemize}

Moreover, an ontology  is seen  as a  very {\em  general} organizational
device: i.e., one  that provides  a classification  system for  whatever
area of  application  the  ontology  is  applied  to. The organizational
resource offered by an ontology has to be {\em re-usable}. But it is  an
open issue  as  to  what  extent  the  kinds of organization listed here
overlap. It cannot be taken for granted that they all refer to the  same
level of abstract description. It can also not be taken for granted that
there is unity  concerning the  {\em tasks}  that are  involved in  such
descriptions. This can be seen in the following statement from Hobbs.
\begin{quote} \small
`Semantics is the attempted specification of the relation between
language and the world. However, this requires a theory of the world.
There is a spectrum of choices one can make in this regard. At one end
of the spectrum --- let's say the right end --- one can adopt the
``correct'' theory of the world, the one given by quantum mechanics and
the other sciences. If one does this, semantics becomes impossible
because it is no less than all of science\ldots There's too much of a
mismatch between the way we view the world and the way the world really
is. At the left end, one can assume a theory of the world that is
isomorphic to the way we talk about it. \ldots Most activity in 
semantics  today is slightly to the right of the extreme left end of
this spectrum. \ldots it fails to move far enough away from language to
represent significant progress towards the right end of the
spectrum.'~\cite[p68]{Hobbs85} 
\end{quote}
It probably does  not make  sense, therefore,  to talk  of a generalized
classification system without first fixing more precisely the nature  of
its intended  function.   A  further  problem  is  that the first of the
desired functions above, organizing world  knowledge, is often taken  to
be {\em  definitional}  for  an  ontology.\footnote{Or the second may be claimed
to be the real task --- however, as Hobbs points out, this actually comes closer to
the first position.}   However, the world --- i.e.,
psychological, logical, or philosophical views of the world --- has  not
proved to be  very constraining  as to  what knowledge  organizations it
requires.  `Ontologies' built on the basis of such constraints are, as
we shall see below, underconstrained and there has accordingly  been no  achievement of the
large scale resources necessary for re-use across {\sc nlp} systems.

The main purpose of this paper is  to add a further round of  discussion
to that concerning the  design and construction  of ontologies for  {\sc
nlp}. The paper is explicitly explorative, building on experience in the
definition and use of such ontologies for text generation. The paper  is
intended to  stimulate  discussion,  rather  than  present solutions ---
although  I  do   conclude  with   suggestions  for   certain  lines  of
theoretically   motivated   methodological   development   for    future
ontologies. The basic path taken in  the paper will be to  differentiate
among the distinct functions  that ontologies may  serve in order  to be
better able to  set out  principles and  constraints for  the design  of
abstract levels of knowledge organization  that can serve as  ontologies
appropriate for {\sc nlp}. Seen in  more detail, the paper is  organized
as follows.

First, I discuss the role of language as a possible motivating force for
designing and populating ontologies. Second, I introduce several of  the
most extensive ontologies that  are currently to  be found in  {\sc nlp}
systems, characterizing  their  precise  function  and motivation within
their respective systems.        Third, I  relate the distinct types  of
ontology discovered to possible  general linguistic theories  that would
support them.          It  is  my  contention  that  many  principles of
organization follow directly  from the  position of  suggested bodies of
information in the  linguistic system  as a  whole and  that recognizing
this allows efforts in the definition and construction of such bodies of
organization to be  directed more  appropriately than  has hitherto been
the case. For  any ontology  that is  proposed, therefore,  it should be
possible to  relate  its  properties  back  to  a  motivating linguistic
theory. I  argue  that  the  evidence  that  we  now  have from the more
extensive attempts  at  ontology  construction  suggests strongly that a
richly {\em stratified} model  of the linguistic  system is required  in
order to achieve  the degree  of constraint  that we  need for attacking
large-scale, re-usable ontology  construction.  Fourth,  I show  how the
ontology of  the  Penman  text  generation  system  ---  that  has  been
developed largely as an instantiation of the highly stratified theory of
systemic-functional  linguistics  ---  already   answers  many  of   the
criticisms that have been raised against other ontologies. I argue  that
although these criticisms  are often  based on  largely {\em  post hoc},
methodological grounds, the vast majority  of them also follow  directly
from the properties of the linguistic system and so could (and  arguably
{\em should}) have been made prior to attempting ontology  construction.
This can be seen  in the properties  of the Penman  ontology, whose very
design avoids  significant  criticisms  levelled  elsewhere.  Finally, I
suggest how ontology design could be improved yet further by taking into
consideration more input from linguistic theory.   The Penman  ontology,
for  example,  is  only  a  partial  instantiation  of  the  theoretical
principles underlying it and it is possible to show that problems  enter
into the  account  precisely  where  the  ontology  falls  short  of the
theoretical specification.

In general,  then,  this  paper  is  intended  not  only  to improve our
understanding of  what  kinds  of  bodies  of  information  can stand as
ontologies of various kinds and how such bodies of information relate to
other resources in  the computational  representation of  the linguistic
system, but also to  make the point  that appropriate views  of the rich
dimensions of organization exhibited by  the linguistic system can  go a
long way to improving  our initial design  specifications for {\sc  nlp}
systems. They should, therefore, always  be considered very early  on in
system construction and computational theory development.

\section{The role of language in  ontology justification}

As mentioned above, the move to consider {\sc nlp} systems that  require
information over  and  beyond  that  attributable  to surface syntax has
raised two problems: how  to organize that information and  how to
relate that information with the less abstract levels of the  linguistic
system. The  first  problem  is  typically  considered in more detail in
approaches where the operation of a  system {\em in some specified
domain}  is the
central goal;  the  second  usually  arises  in systems which attempt to
model the  linguistic  system  itself,  focusing  less  closely  on  the
embedding in any particular specified domain of application.

One common source for knowledge construction and representation that  is
found in approaches to the first  problem is earlier work in  artificial
intelligence ({\sc ai}).        Even  early {\sc ai}  reasoning programs
needed to represent the state of the world in which the programs were to
operate. This has given rise to the areas of {\bf domain modelling}  and
{\bf commonsense  reasoning}  which  are  responsible  for  representing
concrete details of  aspects of  the world.     The enterprise  of world
modelling  clearly  has  many  similarities  with  the  requirements  of
sophisticated {\sc nlp} systems and  there has naturally been  an influx
of techniques and attitudes concerning ontology design from the {\sc ai}
context.

This has  proved  most  successful  in  the  cross-over of techniques of
knowledge representation  in  {\sc  ai}  to  techniques for representing
linguistic  information.           The  similarity  between   structured
inheritance   knowledge   representation   languages   such   as    {\sc
kl-one}~\cite{BrachmanSchmolze85} and its descendents and current  typed
feature logics (e.g.,~\cite{Smolka89,SmolkaAit-Kaci89,Nebel-etal91}) is an active area of research.  A
basic model for the  {\em representation} of  ontologies can now  assume
minimally that a  subsumption lattice  over sorts  is defined,  probably
with some mechanism corresponding to the structured inheritance of  role
information associated with the  sorts, and possibly  additional axioms,
or particular inferences, licenced  by specified combinations  of sorts.
This will be the representational basis for ontologies of all kinds that
I will assume throughout this paper.

In contrast to this  concensus, attempts to  decide exactly {\em  which}
sorts  make  sense  for  an  ontology  based  on  {\sc  ai}   `knowledge
engineering'  principles  have  been   less  successful.  Although   the
effort-intensive  nature  of  domain   modelling  naturally  calls   for
consideration  of  the  re-usability  of  components  of  the  knowledge
represented across distinct  domains, the  ability of  {\sc ai}-centered
approaches to come up with such general organizations has been  limited.
Some of  the  earliest  work  in  this  area was that on `naive physics'
(e.g.,~\cite{Hayes79,Hayes85}):    here  the  aim  was  to  capture  the
underlying `general knowledge' that  people have about  physical objects
and substances in the world; similar investigations are reported in, for
example,~\cite{HobbsMoore85,Hobbs-etal87}, and there are naturally  also
connections to be  drawn with  other work  in semantic  and `conceptual'
representation in {\sc ai}, e.g.,~\cite{SchankAbelson77}.  Further  good
examples of systems that require detailed real `knowledge' in particular
domains are expert systems; here also there is still little shareability
across domain  models.   The  detailed  organization  of  such  systems'
knowledge is  typically  unique  to  particular  application domains and
shows relatively little cross-domain re-usability.

We can  in  part  explain  this  by  considering the relative importance
assigned to the distinct functions that  such domain models in {\sc  ai}
are to fulfill. For example, when constructing a knowledge source  whose
primary function is to  support the particular  inferences that a  given
system needs  to  draw,  it  is  logical  that  the organization of that
knowledge be tailored with this goal in mind.       This usually  leads,
however, to nongeneralizeable representational requirements because  the
inferences that  distinct  systems  are  to  draw  have not been related.   The
relatively small scale  of most  of this  work to  date has  furthermore
limited  the   effectiveness   and   urgency   of   investigations  into
re-usability:  the cost of constructing  domain models from scratch  has
not been  prohibitively  high.   This  cost-equation  quickly
changes  once  more  realistically  sized  bodies  of  information   are
considered.      It quickly  becomes much  more important  that detailed
organizations of  general  knowledge  applicable  to  many  domains  are
available so as to reduce the work involved when moving to new domains.

The  most  extensive  attempt  to  create  a  general  scaffolding   for
representing general, background  knowledge of  the world  based on {\sc
ai} techniques is the {\sc cyc} project~\cite{LenatGuha88}.    The  size
of this project (initial projections were for a base level of 10,000,000
entries) of necessity forces  a sharp awareness  of the need  to have an
organization for  knowledge  that  is  detailed  and  general  enough to
provide sufficient  scaffolding  for  supporting  large-scale  bodies of
information in accessible and usable ways. Without clear principles both
for the organization  of such  knowledge and  for the  selection of  the
information to be represented, the result would be disastrous:    poorly
organized knowledge will  be inadequate  both theoretically,  in that it
fails to capture significant  generalizations, and practically,  in that
it fails to be usable as a resource. The procedure followed in {\sc cyc}
is to divide up types of entities into categories that appear to  behave
differently, i.e., concepts  are classified  according to  the kinds  of
inferences that they  allow to  be drawn  about themselves.  Problematic
here, therefore, is precisely which kinds of inferences are to be  taken
as definitional. This does not appear to have been made explicit and  so
the procedure does not provide a particularly sound methodology.     The
resulting {\em domain-independent}, and hence re-usable, portion of  the
{\sc cyc} ontology is accordingly  not very deep, somewhat  tangled, and
supports limited inferences. It then becomes increasingly necessary
to raise  questions  concerning  the  consistency  of  distinct areas of
knowledge represented and, consequently, how one can use that knowledge.

It needs to be recognized that it is essential to define the  purpose
for which  a  body  of  information  is  to  be  used in order to define
appropriate organizations for that information. As long as the  purposes
are unclear, or too varied,  consistent organizations will be  difficult
to achieve.      The  statement  that  a  general ontology of real-world
knowledge should simply `represent' that knowledge is underspecified.
It
does not provide  sufficient guidance  for finding  useful organizations
for that knowledge. Given that we  need a general organization and  that
that organization will be determined by purpose, we clearly need a  very
general (but still formally  specifiable) task that  requires particular
inferences to be  performed. If  it were  possible to  find such a task,
then it  would  be  possible  to  use  it  as  a guiding methodology for
constructing general organizations of knowledge. Precisely one  such
general task  is,  of  course,  the  expression  of knowledge in natural
language: whatever  the  knowledge  that  is represented, i.e., whatever
domain and however  general/specific, it  should be  possible to express
that knowledge linguistically.\footnote{This is overstated to the extent
that some information/knowledge is often maintained to be  inexpressible
linguistically --- even if this is so, it is still the case that by  far
the widest and most generally applicable form of expression that we know
is language.  In any case, whether or not there exists knowledge that is
inexpressible linguistically will not affect the final outcome of  the
discussion below.} One additional set of constraints that one can  apply
in the construction of organizations  of knowledge that attempt  maximal
applicability across domains is then that offered by language.

This must be specified further. For example, the acceptance of
`ways
of talking'  about  categories  as  evidence  for the existence of those
categories in an ontology is a  very old strategy (e.g., Aristotle)  and
is present even in {\sc cyc}.  This method of justification is, however,
limited to seeing what one can say and still make sense about a category
rather than any  more technical  analysis of  linguistic properties. The
precise  `inferences'  that   are  being   relied  upon   to  shape  the
organization are, therefore,  still not  being given.   Thus, there  are
examples of ontologies that are constructed in {\sc nlp} systems,  where
there is  a  specified  relationship  between  concepts  and  linguistic
expression, but the relationship is  sufficiently non-general so as  not
to provide strong constraints on ontology design.

One    such     case     is     the     ontology     of    {\sc    kbmt}
projects~\cite{CarbonellTomita87}         such          as          {\sc
translator}~\cite{NirenburgRaskin87}.  Work of this  kind seeks a  level
of representation that is minimally different across distinct languages.
Moreover, the value  of organizations  of information  that are relevant
across distinct  domains  is  clearly  recognized and re-usable ontology
portions are actively sought.    However, although the link to  language
ensured by the  machine translation  task increases  the likelihood that
this can be achieved  on a larger  scale, the re-usable  portions of the
ontologies proposed  until  now  remain  small.     This  can in part be
attributed  to  the  fact  that  the  appeal  made  to  language  as   a
constraining force on ontology  design is undervalued.\footnote{This  is
also made problematic by the very multilinguality of possible linguistic
constraints  inherent  in   machine  translation   system  ---   without
appropriate ways of  achieving {\em  linguistic} generalizations  across
languages     (cf.,     e.g.,~\cite{Bateman-etal91-multi-penang}     for
discussion), the application of linguistic constraints is very much more
difficult.}  The  `external-to-language'  attitude  towards  ontological
constructs assumed from {\sc ai} promises to capture abstract models  of
the world (or of conceptions of the  world --- a difference that is  not
criterial at this point) and its organization independent of  particular
languages.   This   appears   a   tempting   direction   for   achieving
interlinguality.    But then we find `motivations' such as the following
for the  categories  that  are  to  be  adopted  within  an interlingual
ontology:

\begin{quote} \small  `Russian
has no word that corresponds exactly  to the English word {\em  afford}
(as in {\em I can't afford X} or I can't afford to Y). In a multilingual
processing enviroment, there might be a concept corresponding to a sense
of the English  word {\em  afford}. A  Russian sentence  {\em Ja ne mogu
sebe etogo pozvolit' (I can't allow myself this)}, uttered in a  context
of acquisition \ldots  should involve  the concept  that represents {\em
afford}. This means that if the units of the representation language are
chosen so that  they are  based on  Russian lexis,  the meaning  of {\em
afford} will be missing. {\bf But this meaning seems sufficiently  basic
to be  included  in  an  ontology.}'  \cite{NirenburgLevin91}  [bold: my
emphasis].  \end{quote}  

It is  clear  that  this  kind  of  argumentation  needs to be sharpened
considerably; it is also clear  that this can only  be done when it  has
been established exactly what function  the `ontology' is to  serve.  In
general,  the  more  detailed  the  linguistic  constraints  adopted  on
ontology design are, the more detailed and {\em explicitly  justifiable}
that ontology design  becomes.\footnote{This was  also one  result of an
extensive    study    of     proposed    ontologies     reported    upon
in~\cite{SkuceMonarch90}. Although  there  has  also  been  at least one
example of  development  that  has  attempted  movement  in the opposite
direction.  The {\em  abstraction structure}  of BBN's  natural language
and understanding  project  {\sc  janus}  was  redesigned  away  from  a
linguistically oriented description  in order  to find  a `more  general
ontological style'~\cite[200]{Weischedel89}  that  was  not  so strongly
connected with  the  linguistic  realization  of  the  concepts defined.
However, this very move was probably one contributing factor to the less
than successful  outcome  of  the  subsequent  attempt  to  use the {\em
Longman Dictionary of Contempory  English} as the  basis for defining  a
domain-independent taxonomy  for  {\sc janus}~\cite{ReinhardtWhipple88}.
The most significant generalizations that would have helped organize the
taxonomy for the purposes of  natural language processing had probably
already been lost.}  However,  the  relationship  between ontologies and {\sc
nlp} is interestingly reflexive.\footnote{Or even circular:  as I  shall
mention below.}  Ontologies  appear  necessary  for  the organization of
knowledge appropriately for use by {\sc nlp} systems, and simultaneously
the explicitness of  the necessary  inferences that  constitute an  {\sc
nlp} system provide  an until  now unrivalled  source of  constraint for
deciding on ontology designs.

This connection is described well  in the following citation  from Ewald
Lang:  \begin{quote} \small  `\ldots the  structure of  language plays a
dual role.   It  is, properly  allocated to  the parsing  and generating
components, a constitutive part  of the object  to be modeled  (that is,
the  system  which  is   to  integrate  linguistic   and  non-linguistic
knowledge). But at the same time it is also part of the device by  means
of which this object is accessed, that is, the categorization of lexical
items into nouns, verbs, etc.,  provides an apparently natural  grid for
establishing corresponding sorts of entities in the ontology, which,  by
definition, is to represent non-linguistic common sense knowledge. Given
this, the risk of confusing linguistic and non-linguistic categories  is
latent; moreover,  it  is  practically  unavoidable  as  long  as we are
confined (or confine  ourselves) to  looking at  common sense  knowledge
through the window of language only,  i.e., without a chance to  draw on
independent evidence from  non-linguistic (say,  visual or  kinasthetic)
ways  of  accessing   the  structure   and  contents   of  common  sense
knowledge.'~\cite[p464]{Lang91}  \end{quote}   Thus,   while  linguistic
patterns are probably the richest source of organizational criteria that
are  available  to   ontology  design,   their  use   is  certainly  not
unproblematic.  Consequences  of  this  can  be  seen  in  the fact that
although the majority of recent  and currently planned natural  language
processing systems recognize  the necessity  of some  level of  abstract
`semantic' organization similar to an ontology that classifies knowledge
explicitly    according    to    its    possibility    for    linguistic
expression,\footnote{Including, for example:    the Functional  Sentence
Structure of {\sc xtra}:~\cite{Allgayer-etal89}; \cite{Dahlgren-etal89};
\cite{Emele89}; the  {\sc  polygloss}  project:     \cite{Emele-etal90};
certain  of   the   domain   and   text   structure   objects   of  {\sc
spokesman}:~\cite{Meteer89}; {\sc translator}:  \cite{Nirenburg-etal87};
the Semantic Relations of  {\sc eurotra-d}:  \cite{Steiner-etal87};  the
{\sc janus} project:~\cite{Weischedel89}; and  the ontological types  of
the  {\sc   acord}   project:~\cite{Moens-etal89}.             Moreover,
ontology-like organizations of informations have also been found  useful
for          parsing           applications           by,          e.g.,
~\cite{Calder-etal89,ChenCha88,Hinrichs-etal87,Zajac89}.  There  are  no
doubt many other places where this kind of construct now appears.}  very
few have  achieved  ontologies  of  any  size  and {\em motivations} for
inclusion of particular concepts  and distinctions in  ontologies remain
limited or underspecified. Thus, the decision to use linguistic evidence
by itself is still, unless further restricted, underspecified and leaves
open a range of positions. These give rise to differing  functionalities
that the  ontologies  are  to  serve,  which  hence  impacts on ontology
design. The positions and functionalities need to be characterized more
precisely and this I attempt in the  following  section.

\section{Three kinds of ontologies}

Although I have  concentrated until  now on  preliminaries to  the first
problem area  mentioned  in  the  introduction  --- how knowledge of the
world is to be represented --- the apparent value of applying linguistic
constraints to this task  renders the second  problem area ---  how that
knowledge is related to language --- crucial. If the ontology cannot  be
related to  language  in  an  explicit,  formalized  fashion,  then  the
structures (and functions) of language  will be prevented from  having a
direct constraining influence on what gets represented in the  ontology,
what not, and how the entire ontology is to be organized.

\begin{figure*}
\rule{\textwidth}{0.2mm}
\epsfig{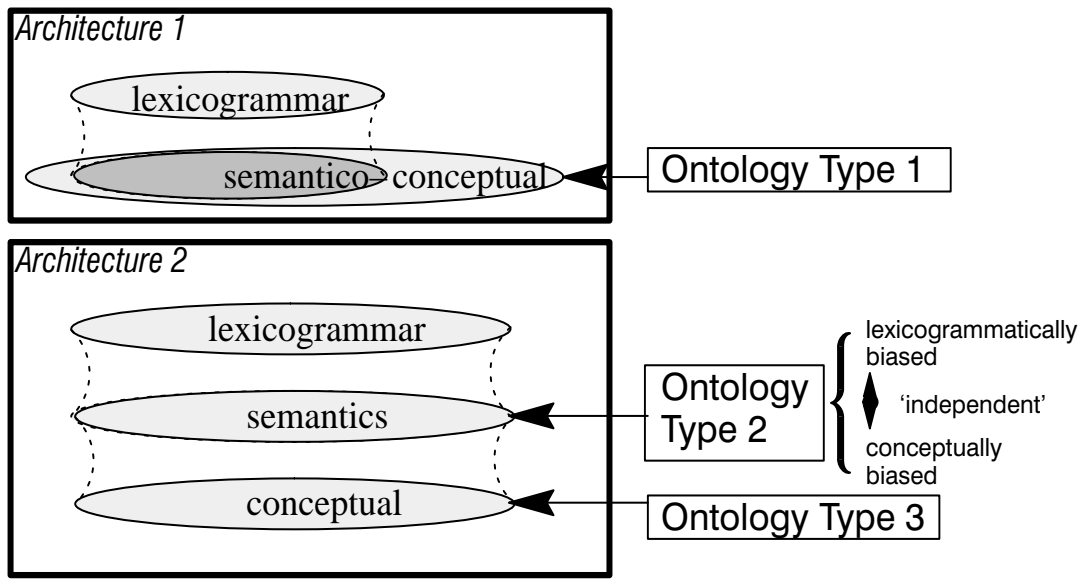}{90}{150}
\caption{Three kinds of ontology in \sc nlp}
\label{3-ontologies}
\rule{\textwidth}{0.2mm}
\end{figure*}

There are at  least two  theoretically distinct  standpoints from  which
this second problem area has been addressed in {\sc nlp} systems.    One
possibility is to  assume that  real-world domain  knowledge is  more or
less directly linked to grammatical and lexical forms of expression. The
organization of the world knowledge ontology should then, ideally,  also
be supportive of the use of that knowledge for linguistic expression  or
for interpreting  linguistic  distinctions:   the  problem  of  relating
knowledge to  language  is  thus  subordinated  to  the  world knowledge
ontology design. A  second  possibility  is  to  assume  that  the
relationship  between
real-world domain knowledge  and grammar  and lexis  is itself complexly
structured. This structuring may lean  for its organization towards  the
world knowledge ontology, in which case this would blend into the  first
possibility, or towards the  grammar and lexicon, or  alternatively
could rely  on  its  own  principles  of  organization.   Each  of these
variants has been adopted in some  system where a concrete ontology  has
been attempted. This gives rise to three distinct kinds of ontology that
can be found in {\sc nlp} work.
An ontology can be
\begin{itemize}
\item an  abstract  semantico-conceptual  representation  of  real-world
knowledge that also  functions as  a semantics  for use  of grammar  and
lexis --- this type I will term a {\em mixed} ontology: $O_m$; 

\item an abstract organization underlying our use of
grammar and lexis that is separate from the conceptual, world
knowlege ontology, but which acts as an interface between grammar and
lexis and that ontology --- this type I will term an {\em interface} ontology: $O_i$; 

\item an abstract organization of real-world knowledge (commonsense or
otherwise) that is essentially {\em non-linguistic} --- this type
I will term a {\em conceptual} ontology: $O_c$.
\end{itemize}
The relationship involved here, their embedding in general
architectures, and the subtypes of interface ontologies mentioned above
are depicted graphically in Figure~\ref{3-ontologies}. 

\subsection{Conceptual ontologies}

Most of the  {\sc ai}  designed ontologies  --- including  those of {\sc
cyc},          {\sc          tacitus}~\cite{Hobbs-etal87},          {\sc
janus}~\cite{Weischedel89},        `the         naive         semantics'
of~\cite{Dahlgren-etal89}, and even  some aspects  of the  {\sc kbmt}
ontology,   e.g.,~\cite{NirenburgRaskin87,NirenburgLevin91}   ---    are
attempts to  construct  ontologies  of  the  third  type: pure maximally
language independent ontologies reflecting  the structure of  the world.
I have  already  discussed  some  of  the difficulties of designing such
ontologies  without  building  up   through  an  account   of  language.
Psychological research might offer another  source of evidence for  such
ontologies; as  would  detailed  sociological  work  on  the commonsense
world. It is, however, unclear whether any such methodology will be able
to avoid the relationship to language  observed above and so I  will now
concentrate on  ontologies  which  are  at  least intended to be related
explicitly to language. 

\subsection{Mixed ontologies}

An example of a mixed ontology --- i.e., one where there is no extensive
treatment of  the  relation  between  the  world  knowledge ontology and
grammar and lexis maintained  separate  to the ontology  itself --- is 
the approach
taken   in   the    {\sc   lilog}    natural   language    understanding
project~\cite{HerzogRollinger91}; details of the ontology are given  in,
for example,~\cite{Klose-vonLuck91,Pirlein91,Klose-etal91}, and  details
of the relation between linguistic forms and conceptual  representations
are given  in,  for  example,~\cite{Gust91,Bosch91}.   It  may  at first
glance appear strange to classify  {\sc lilog} here, since  the approach
to the  relation  between  linguistic  form  and  world  knowledge draws
heavily  on~\cite{Bierwisch83}'s   theory   of   semantics   where,   to
cite~\cite{Gust91}'s statement  of  Bierwisch's  position:       `\ldots
semantic  forms  and  conceptual  structures  belong  to  different  and
strictly discriminated levels'. \cite{Bosch91} also makes it very  clear
that he holds this distinction to be crucial  for making
progress
in semantics and knowledge  representation. However, when  the modelling
of the approach  is examined,  we find  that this  distinction of levels
comes under attack.        For example,  both semantic forms, which  are
derivable from the lexicon and from grammatical analysis, and conceptual
forms are represented in a single language (the semantic language  being
a subset  of  the  conceptual  language:~\cite[p248]{Bosch91})  and  are
freely combinable; moreover~\cite[p133]{Gust91}  maintains  that: `there are
continuous variations between semantic forms and conceptual structures.'
This gives rise to lexical entries which directly contain categories  of
an ontology which also contains categories of real-world knowledge.  The
relation between conceptual knowledge  and grammatical and  lexical form
is thus handled by logically manipulating categories from a single
ontology until categories are found that possess links to grammatical or
lexical entries --- this is precisely the architecture consistent with a
mixed ontology as shown in Figure~\ref{3-ontologies}.

An illustration of the nonseparation of of `linguistic'
information and
`conceptual' information typical of ontologies of this type can
be seen in the following
taken  from~\cite[p251]{Bosch91}. In order to find the interpretation in
context of the lexeme ``school'' as it is used, arguably
differently, in examples such as:
\begin{quote} \small
a. The school made a major donation.\\
b. The school has a flat roof.
\end{quote}
A general `lexical semantic entry' for the lexeme is retrieved thus:
\begin{quote} \small \tt
SEM(``school'') = $\lambda $X [PURPOSE X W] \\
\rm where\\
\tt W = PROCESSES\_OF\_LEARNING\_AND\_TEACHING
\end{quote}
This is then interpreted by applying a given `contextualizing function'
selected depending on the basis of the semantic interpretation of the
predicate in the lexicogrammatical representation. Those for the example
sentences would be:
\begin{quote} \small \tt
a. $\lambda $X [INSTITUTION X \& SEM X]\\
b. $\lambda $X [BUILDING X \& SEM X]
\end{quote}
Combining the semantic entry and the contextualizing function gives the
required `conceptual' concept that is the referent of the lexeme in
context --- i.e., that ``school'' is interpreted as either an
institution or a building. All of the undefined predicates found in
these logical expressions (e.g., {\sc institution, purpose},
etc.) are sorts defined in the ontology. A direct link is therefore
constructed from lexicogrammatical information and chunks of information
appropriate for the conceptual level of organization. As we shall see in
Section~\ref{ling-motivation}, this direct linking is a common property
of {\sc nlp} systems based on the common notion of `semantics' and
arises out of a view of the linguistic system that collapses together
several important distinctions.

\subsection{Interface ontologies}
\label{ontology-examples}

The second  and  third  types  of  ontology  --- the interface and
conceptual types --- usually occur, at least theoretically, in the  same
architecture. Although it  is also  the case  that some  systems address
themselves  to  the  organization  of  the  interface  ontology  without
specifying how the conceptual ontology  will look. This latter  position
is common for  systems that  are intended  as general  purpose {\sc nlp}
systems    re-usable  across  different  domains   and
applications.   Examples  of  such  systems  include  both  parsers  and
generators such as  the Penman  system~\cite{MannMatthiessen85,Penman89}
and Mumble-86~\cite{Meteer-etal87}. Here the problem of how to  organize
the interface with external  applications, where those  applications are
not known in advance, has  naturally focused attention on  organizations
of information appropriate for  interfacing.       The approach  to this
developed within the Penman  project in terms  of the {\bf  Upper Model}
has become more or less typical of how this is achieved --- although  to
what extent this architecture has arisen independently across systems is
unclear. The initial formulation of the Upper Model was based on work by
M.A.K.      Halliday  ~\cite{Halliday82},  William  Mann  and  Christian
Matthiessen.\footnote{The development of the Upper Model ontology,  from
its inception as the {\em Upper structure} of the {\sc janus} project of
ISI and BBN, up to its inclusion as a standard component of the  current
Penman text  generation  system  is  covered  by  the following research
reports:~\cite{Mann85-janus2,Mann-etal85-janus,MooreArens85,UM-89}.  The
first detailed theoretical precursor to the ontology was set out in 1985
by  Halliday  and   Matthiessen  as   a  general   organization  for  an
experiential semantics: this was  called the {\bf  Bloomington Lattice}.
The subsequent development of the Upper Model has deviated somewhat from
the purely linguistically motivated work; this will be discussed in more
detail below.}

The general statement of the interface problem for {\sc nlp}
systems is that machine-internal
information needs  to  be  related  to  strategies  for  expressing that
information in some  natural language.         This could  be done  in a
domain-specific way by  coding how  the application  domain requires its
information to appear.      This  is clearly problematic,  however:   it
requires detailed knowledge on  the part of  the system builder  both of
how the generator controls its output forms and the kinds of information
that the application domain contains.    A more general solution to  the
problem of  defining  a  mapping  between  knowledge  and its linguistic
expression is to provide a classification of any particular instances of
facts, states of affairs, situations, etc.  that occur in terms of a set
of  general  objects  and  relations  of  specified  types  that  behave
systematically with respect to  their possible linguistic  realizations.
This classification has itself  many of the  properties of an  ontology,
e.g., it is a hierachical organization  of sorts and roles ---  although
by virtue of its motivation in  linguistic realization, it must be  seen
as a strictly linguistically  motivated ontology. Examples  here include
aspects of \cite{Meteer89}'s  {\em Text  Structure Objects}  in the {\sc
spokesman} text generator:
\begin{quote} \small
`[i]t  is important to remember  that Text
Structure objects reflect the semantic  type of the {\em  expression} of
the information in  an object,  not some  intrinsic type  of the  object
itself.'\cite[p21]{Meteer89};
\end{quote}
also the ontology of the {\sc acord} system:
\begin{quote} \small
`\ldots the aim of the sort system is not to reflect the characteristics
of real world objects and events referred to by linguistic  expressions,
but  rather  to  systematize  the  ontological  structure  evidenced  by
linguistic expressions'~\cite[p178]{Moens-etal89};
\end{quote}
and,  of course,  the
Upper Model of  the Penman  system that  I will  describe in more detail
below.

The position  that  such  an  interface  ontology  holds between surface
details of a language and more abstract knowledge is, however, an uneasy
one.  As suggested  above, it  is possible  to differentiate  among such
ontologies according to whether  they orientate themselves  more towards
less abstract or  towards more  abstract levels  of representation. This
brings with it two potential problems in ontology design:
\begin{itemize}
\item the ontology can be too {\em shallow}, in that it's categories are
a too direct recoding of linguistic distinctions that do not achieve a
qualitative increase in abstraction;

\item the  ontology  can  be  too  {\em  deep},  in that it is no longer
possible  to  draw  any  formally  specifiable  connection  between  the
constructs posited and the linguistic evidence taken as motivating them.
\end{itemize}
Both extreme situations occur and both reduce the value of the ontology
as an effective interface medium.  The former  problem will  be
accompanied  by an  increased
difficulty in linking the ontology to information of particular  domains
--- regardless of whether this  information is considered as  a separate
kind of information  or as  more specific  details of  the same  kind of
information; and  the  second  problem  will  be  accompanied both by an
increased difficulty  in  linking  with  grammar  and  lexis  and by the
problems induced by poorer linguistic constraints mentioned above.   The
latter situation then often places  a heavier reliance on  `internal' or
formal         constraints         on         organization         (cf.,
e.g.,~\cite{Weischedel89,Horacek89} and what~\cite[p468]{Lang91} terms
`sortal' restictions) which, while important, do  not  provide sufficient
grounds for deducing very much detailed actual {\em content} by themselves. 

\subsubsection{Interface ontologies that are not abstract enough}

Interface ontologies exhibiting the former  problem are very common  and
so it  is  worthwhile  giving  a  slightly  more detailed example of the
problems that arise.  One  such ontology is  that constituted by   the
semantic relations used within the german component of the {\sc eurotra}
project   \cite{Steiner87,Steiner-etal87,SteinerReuther89}.        These
relations are  a  further  development  of  earlier  work by Fawcett ---
particularly   his    work    on    {\em    transitivity}   in   English
(e.g.,~\cite{Fawcett87}).  Fawcett  proposes  a  semantically  motivated
taxonomy  of   process   types,   analogously   to  the  approach  taken
in~\cite{IFG} but differing  in the  actual categories  adopted.    Each
process type has some distinctive  set of possible participants  --- the
approach thus differs from early accounts of semantic participants, such
as Case Grammar \cite{Fillmore68},  where the participant  relationships
were  often  defined  separately  from  the  processes  in  which   they
participate, and further articulates conceptions of `thematic' relations
such   as   those    found   in    Lexical-functional   grammar    (cf.,
\cite{HaleKeyser86,Levin87})     and     Government     and      Binding
theory~\cite{Jackendoff87}. The {\sc eurotra-d} work has made  refinements
to the  proposed  taxonomy  on  the  basis  of  multi-lingual  evidence,
particularly from German, so as to provide explicit {\em syntactic} tests
for the assignment of  processes to each  of the various  process types.
It is then explicitly stated that the resulting process types  described
are {\em no  longer} primarily  semantic since  their classification  is
based exclusively on differentiation by syntactic criteria.   Therefore,
although this has  produced a  framework within  which processes  can be
classified according  to  the  given  taxonomy  with  a  high  degree of
inter-coder consistency,  which  is  an  important  criterion  in  large
distributed projects such as {\sc eurotra}, its effectiveness as a  step
towards a  higher  level  of  abstract  information has been restricted.
This can  be  seen  in  the  following example of process classification
given in \cite{SteinerReuther89}. For the clause 
\begin{quote} \small
That she  gave  no  answer  means  that  she  agrees  with the proposal.
\end{quote} 
both subject and object are realized by {\em that}-clauses
and the  only  possible  classification according to the syntactic tests is 
then  one  of a {\em mental
process} with two {\em  phenomena}.   However, semantically  the process
also has strong elements  of a {\em  relation} between the  propositions
involved.   Similar  examples  in  German  are the following: 
the verb {\em retten \ldots vor}:  
\begin{quote} \small  
Da\ss  \ er  gut schwimmen  konnte,
rettete ihn vor dem Ertrinken.\\ 
{\em That he could swim well saved  him
from drowning.} \end{quote} 
again resembles a relational
process but  has  to  be  assigned  to  mental according to the criteria
formulated; the 
process `reden' ({\em to speak, talk}), which would intuitively seem  to
be some kind of communication  verb, cannot enter into  constructions of
the form:   
\begin{center} \small *  Peter redet:   Karl kommt  morgen\\
{\em Peter speaks: Karl is coming tomorrow} 
\end{center}
and  so does  not receive  a communication  verb
classification: and the  form:  
\begin{center}
\small 
*  Peter redet, da\ss \ Karl kommt \\
{\em Peter speaks that Karl is coming} 
\end{center}
cannot occur so it may
not even receive a mental reading ---  the only acceptable forms  possible,
e.g.:   
\begin{center} \small
Peter redet Unsinn\\
{\em Peter speaks nonsense}\\
Peter redet mit Paul
{\em Peter speaks with Paul}
\end{center}  
require   an   action   classification,   just   as   the
corresponding English processes  would.  These  problems provide  evidence
that the syntactic tests need to  be made more subtle or  more elaborate
in order to be able to reveal semantic distinctions more
reliably.
In addition, there is no account suggested of how this level of
representation can link to more abstract levels of representation such
as a conceptual ontology.

A similar case of this probably contributes to some of the  difficulties
that arise with the use of `Lexical Semantic Structures' ({\sc lss}) and
~\cite{Jackendoff83}'s `Lexical Conceptual  Structures' ({\sc  lcs}) for
translation --- the former as described by~\cite{Dorr91}, the  latter
by~\cite{NirenburgLevin91}.    Both structures are  tightly
bound to  possible  surface  forms  by  formally  specified {\em linking
rules} (e.g.,~\cite{Levin87}).   These rules partition the {\sc lss}  or
{\sc lcs} into classes reflecting the different realizational  behaviour
of their categories.   Although  it is also  then sometimes possible  to
assign to these  classes particular  `semantic' features
% ~\cite{Levin89},
this has still not yet been found to be sufficiently abstract to support
a motivated construction of the corresponding conceptual ontology --- as
the example of  the motivation  for including  the concept  {\em afford}
for Russian that I  cited  above  shows.        The  final  selection 
of conceptual
ontological sorts in this  case then shows  similarities both with  that
described for  {\sc  lilog}:   i.e.,  by  applying a mixture of lexical,
grammatical, and domain knowledge criteria,  and with the pure  {\sc ai}
techniques of  {\sc  cyc}  and  others.  In  the longer term, therefore,
similar problems will occur.

\label{um-intro}
As a final example of the problems  of  lack  of  abstraction, I will
mention some that have arisen in our development and use of the the Penman
Upper Model. The Upper  Model, for reasons  that I
will  describe below,
does succeed in being more abstract than the semantic relations adopted,
for example, within {\sc  eurotra-d}. The  organization of an  Upper
Model achieves  greater  {\em  semantic} coherence, grouping together
distinctions that  may  be  used  by  a  variety of distinct grammatical
resources in a grammar.  For example, the relationships between  process
and participants may  drive the  organization of  clauses, but  they may
equally drive the organization of head and modifiers in nominal  groups.
The nature of  the process-participant  relationships is  not, arguably,
altered by  their  realizational  form.  Upper Model 
generalizations might then express  the  commonality  that unites the
following area of variation:
\begin{center} \small
       A shoots B\\
       B was shot at T\\
       the shooting of B by A\\
       A's shooting of B\\
       B's shooting\\
       the shooting at P\\
       the P shooting\\
       the T shooting\\
       {\em etc}.
\end{center}

under a single specification:\footnote{Although the actual
representation used in the Upper Model
reifies both predicates and the relations holding between
predicates and their arguments; cf.~\cite{Mann-etal85-janus,Hobbs-etal87,UM-89}.}
\begin{center}
process: shoot(murderer:A,murdered:B,time:T,place:P).
\end{center}
Categories in the Upper Model  are then capturing
generalizations which are not appropriately expressed within the
grammar.  A further example drawn from the 1989 version of the Upper
Model is the possible grammatical realizations of the concept of {\em
generalized
possession}.  This concept should be seen as being realized by
possible selections from all the grammatical systems to do with
`possession'. Thus the semantics of the following forms
all make reference to this single Upper Model concept.
{\small \begin{center}
the door's handle\\
the handle of the door\\
the handle that the door has\\
the door handle\\
the handle is part of the door\\
{\em etc}.
\end{center}}
For a more extensive sets of examples of the lexico-grammatical
variation that the Upper Model is intended to support,
see~\cite{Bateman89-IAI,Bateman90-coling}.

This means that the Upper Model does succeed in achieving a sufficiently
high degree of abstraction as to  be useful as an interface  medium. This
increase in abstraction also makes the ontology better suited to  linking
with more abstract  levels of  information.\footnote{So much  so that it
has sometimes  been  our  experience  that  the  domain  model  of  some
application domains  has  been  altered  in  the light of the consistent
organization that the Upper Model brings to bear.} The Penman system has
been successfully interfaced  with a  number of  applications --- mostly
expert systems, but  also text  planners ---  where domain  knowledge is
represented. It is  then an  example of  an ontology  that mediates  the
relationship between lexico-grammar and  world knowledge without  losing
the necessary formal connection with the grammar and lexis. Moreover, it
moves beyond problems such  as that recognized  for the {\sc  eurotra-d}
classification of process types that\footnote{This problem arose while
attempting  to interface the level of
input specification for an existing  generator of German ({\sc  semsyn},
cf.:   \cite{Roesner88})  with  the  
{\sc eurotra-d} semantic
relations.}
\begin{quote}  \small  
`The  classification  system  proposed  by  {\sc
eurotra-d} proceeds  in  a  strictly  syntactic  way.  \ldots  From  the
standpoint of  generation  this  solution  is  problematic:  it would be
preferable to  have  a  semantic  classification that generalizes across
such   surface    syntactic    subtleties.''    \cite[p158]{Heid-etal88}
\end{quote}
To the extent that it is successful, this is precisely what the Upper
Model provides. It achieves this by being based very closely not only on  a
particular, specified  grammar  ---  no  concepts  are admitted into the
Upper Model, for example, {\em unless they have a direct and specifiable
consequence for the operation of the grammar} --- but also on a  grammar
which is {\em itself} already more abstract than a constituency grammar.
I shall describe this in more detail below. 

The Upper Model  thus stands  as a  significant step  forward in dealing
with the problem  of interfacing  with a  general {\sc  nlp} system. The
Upper Model decomposes the mapping  problem inherent in relating  domain
knowledge  with   its   possibilities   for   linguistic  expression  by
establishing a  level  of  {\em  linguistically  motivated\/}  knowledge
organization specifically  constructed  as  a  reponse  to  the  task of
constraining linguistic realizations.  While it may not be reasonable to
insist that {\em application domains} organize their knowledge in  terms
that respect  linguistic  realizations  ---  as  this  may  not  provide
suitable organizations for, e.g., domain-internal reasoning --- we  have
found that  it  {\em  is}  reasonable,  indeed  essential,  that  domain
knowledge be so organized if it is also to support expression in natural
language relying on general natural language processing capabilities.

The general types constructed within the Upper Model necessarily respect
generalizations concerning how distinct semantic types can be  realized.
We then achieve the necessary  link between particular domain  knowledge
and the  Upper  model  by  having  an  application  {\em  classify}  its
knowledge organization in terms of the general semantic categories  that
the Upper Model provides. This should not require any expertise in
grammar
or in the  mapping between  Upper Model  and grammar.    An application
needs only to concern  itself with the  `meaning' of its  own knowledge,
and not  with  fine  details  of  linguistic  form.  This classification
functions solely  as  an  interface  between  domain knowledge and Upper
Model; it does  not interfere  with domain-internal  organization.   The
text generation system  is then  responsible for  realizing the semantic
types  of   the   level   of   meaning   with   appropriate  grammatical
forms.\footnote{This is  handled  in  the  {\sc  penman}  system  by the
grammar's  {\em  inquiry  semantics},  which  has  been  described   and
illustrated                    extensively                     elsewhere
(cf.,~\cite{Penman89}) and see
Section~\ref{ling-motivation} below.} Further,   when   this
classification has been established for a given application, application
concepts  can  be  used  freely  in  input  specifications  since  their
possibilities for linguistic realization are then known. Interfacing with
such a system is thus radically simplified on two counts:
\begin{itemize}
\item  much of the information specific to
language processing is factored out of the input specifications required
and into the relationship between Upper Model and linguistic resources;

\item the  need  for  domain-specific  linguistic  processing  rules  is
greatly reduced  since  the  Upper  Model provides a domain-independent,
general and  reusable  conceptual  organization  that  may  be  used  to
classify all domain-specific knowledge when linguistic processing is  to
be performed.  
\end{itemize}

An example of the simplification that use of the Upper Model offers  for
a text generation system interface  language can be seen  by contrasting
the  input  specification  required   for  generators  that   work  with
realization classes that are less abstract than those of the Upper Model
---   such   as,   e.g.,   {\sc   mumble-86}~\cite{Meteer-etal87},    or
unification-based  frameworks,  such  as~\cite{McKeownParis87}  and  the
Lexical Functional  Grammar  (LFG)  approach of~\cite{MommaDoerre87} ---
with the  input  required  for  Penman.\footnote{Note  that  this is not
intended to single  out these  approaches at  all, the  problem is quite
general  and  occurs  whenever  there  is  no  ontology  available   for
organizing information at a more abstract level than that imposed by the
grammar. Further, as already noted, most current {\sc nlp}  developments
are moving in a  direction analogous to  that taken in  our work on  the
Upper Model.}  Figure~\ref{mumble-comp}  shows  corresponding inputs for
the generation  of  the  simple  clause:          {\em Fluffy is chasing
little mice}.      The  appropriate  classification  of domain knowledge
concepts such as {\em chase, dog,  mouse}, and {\em little} in  terms of
the general  semantic  types  of  the  Upper  Model  (in this case, {\em
directed-action,  object,  object,}  and  {\em  size}  respectively  ---
cf.~\cite{UM-89}) automatically  provides  information  about  syntactic
realization that needs to  be explicitly stated  in the {\sc  mumble-86}
input (e.g.,  {\tt  S-V-O\_two-explicit-args, np-common-noun,}\linebreak
{\tt restrictive-modifier, adjective}).              Thus, for  example,
the classification of a  concept {\em mouse}  as an {\em  object} in the
Upper Model is sufficient for the grammar to consider a realization such
as, in {\sc mumble-86} terms, a {\tt general-np} with a particular  {\tt
np-common-noun}  and   {\tt   accessories}   of   {\tt  gender  neuter}.
Similarly, the classification of {\em chase} as a {\em  directed-action}
opens up linguistic realization  possibilities including clauses  with a
certain class of transitive  verbs and characteristic  possibilities for
participants,  corresponding  nominalizations,  etc.     Such  low-level
syntactic   information   is    redundant   for    the   {\sc    penman}
input.\footnote{Moreover, when additional information is required,  that
information is supplied in {\em semantic} terms rather than in terms  of
morphosyntactic labeling such as {\tt  :number plural} --- in  this case
this is represented in inquiry  semantics by the inquiry  response pairs
\{:multiplicity-q multiple\} and \{:singularity-q nonsingular\}.    This
is also  the  case  for  `tense'  but  I  have abbreviated the semantic
specification here.    For  descriptions  of  all  these distinctions in
detail, see  the  {\sc  penman} documentation~\cite{Penman89}.} Similar,
illustrative inputs  forms  can  easily  be  imagined for other types of
syntactically oriented grammar and lexis components.

\begin{figure*}
\rule{\textwidth}{0.5mm}
{\small \begin{verbatim}
(general-clause
  :head (CHASES/S-V-O_two-explicit-args
          (general-np
            :head (np-proper-name "Fluffy")
            :accessories (:number singular
                          :gender masculine
                          :person third
                          :determiner-policy no-determiner))
          (general-np
            :head (np-common-noun "mouse")
            :accessories (:number plural
                          :gender neuter
                          :person third
                          :determiner-policy initially indefinite)
            :further-specifications
              ((:attachment-function restrictive-modifier
                :specification (predication-to-be *self*
                                   (adjective "little"))) )) )
  :accessories (:tense-modal present :progressive
                :unmarked) )
\end{verbatim}
% \vspace{-0.3in}
\begin{center}
{\small Input to {\sc mumble-86} for the clause: \em Fluffy is chasing
little mice}\\
from: ~\cite{Meteer-etal87}\\
\rule{3in}{0.1mm}
\end{center}
\begin{verbatim}
(e / chase
   :actor (e / dog :name Fluffy)
   :actee (m / mouse
               :size-ascription (s / little)
               :multiplicity-q multiple :singularity-q nonsingular)
   :tense present-progressive)
\end{verbatim}
\vspace{-0.3in}
\begin{center}
{\small Corresponding input to \sc penman}\\
\end{center}}
\vspace{-0.3in}
\caption{Comparison of input requirements for \sc mumble-86
\rm and \sc penman \label{mumble-comp}} 
\rule{\textwidth}{0.5mm}
\end{figure*}

\begin{figure*}
\rule{\textwidth}{0.5mm}
%\input{reusuable-um}
%\centerline{\box\graph}
\epsfig{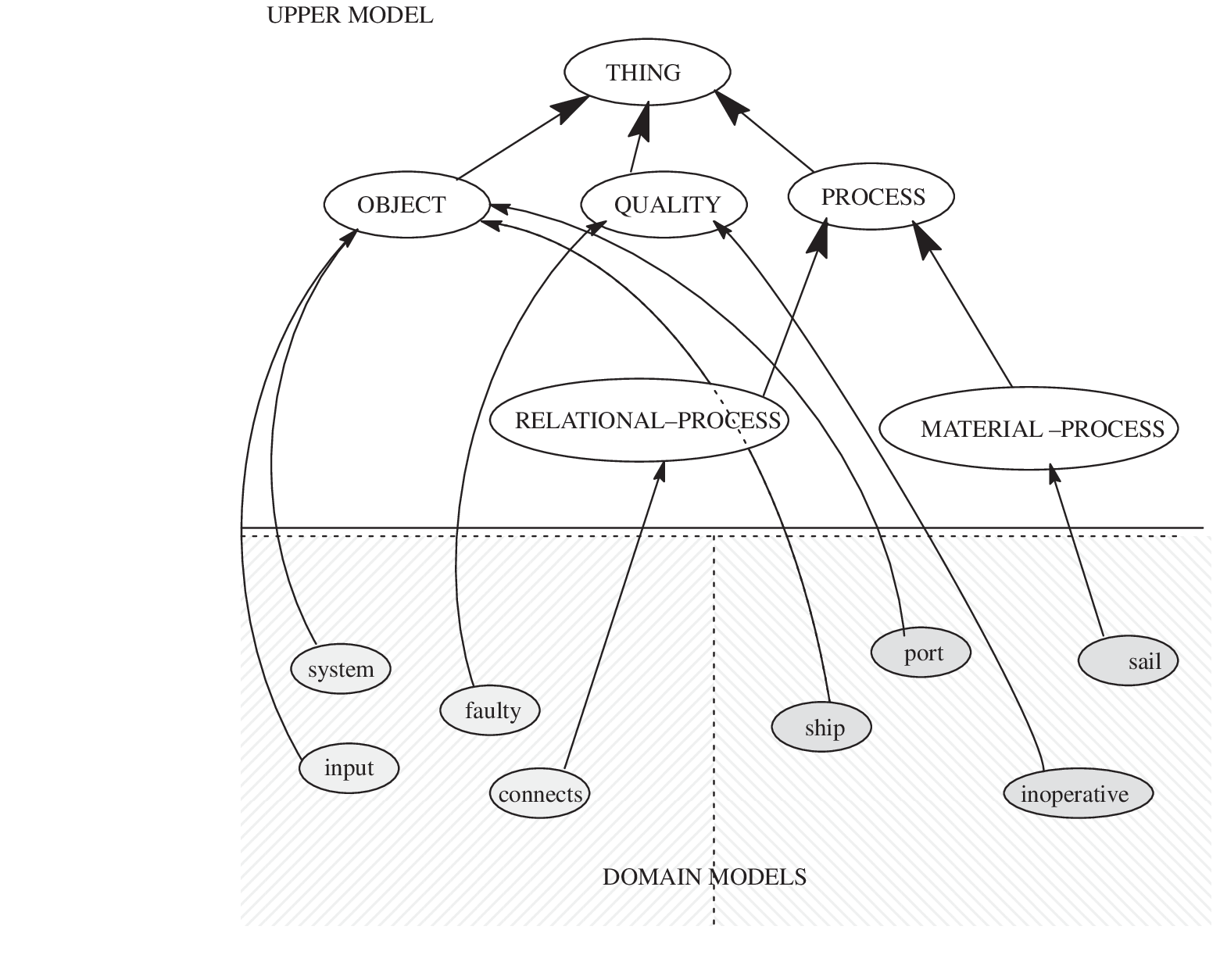}{150}{150}
\caption{Upper Model organization reuse with differing domains}
\label{reusable-um}
\rule{\textwidth}{0.5mm}
\end{figure*}

The further {\em domain-independence} of the Upper Model is shown in the
following example  of  text  generation  control.    Consider two rather
different domains: a  navy database  of ships  and an  expert system for
digital circuit diagnosis.\footnote{These are, in fact, two domains with
which we have had  experience generating texts  using the Upper  Model.}
The navy data  base contains  information concerning  ships, submarines,
ports, geographical  regions,  etc.  and  the  kinds  of activities that
ships,  submarines,  etc.  can  take  part  in.    The  digital  circuit
diagnosis expert  system  contains  information  about  subcomponents of
digital circuits, the kinds of connections between those  subcomponents,
their possible functions, etc. A typical sentence from each domain might
be:  
\begin{quote} 
{\em circuit domain:} The faulty system is  connected
to the input.\\  
{\em navy  domain:} The  ship which  was inoperative  is
sailing to Sasebo. 
\end{quote} 
The input specifications for both of these
sentences   are   shown    in   Figure~\ref{navy-circuit-spls}.    These
specifications freely  intermix  Upper  Model  roles and concepts (e.g.,
{\em domain,  range,  property-ascription})  and  the  respective domain
roles and concepts (e.g., {\em system, faulty, input, destination, sail,
ship, inoperative}).  Both  forms  are  rendered  interpretable  by  the
subordination of the domain concepts to the single generalized hierarchy
of   the   Upper   Model.    This   is   illustrated    graphically   in
Figure~\ref{reusable-um}. Here we see the single hierarchy of the  Upper
Model being  used  to  subordinate  concepts  from  the two domains. The
domain concept {\tt system}, for  example, is subordinated to  the Upper
Model concept {\em  object}, domain  concept {\tt  inoperative} to Upper
Model concept {\em quality}, etc. By virtue of these subordinations, the
grammar  and  semantics  of  the  generator  can  interpret  the   input
specifications in order to produce appropriate linguistic  realizations:
the Upper  Model  concept  {\em  object}  licenses  a  particular set of
realizations, as do the concepts {\em quality}, {\em  material-process},
etc.\footnote{For further discussion of this simplification in the
semantic   input    specification    for    the    sentence   
generator,  see~\cite{Bateman90-nlgw5}.}

\begin{figure*}
\rule{\textwidth}{0.5mm}
\begin{verbatim}
(v1 / connects
    :domain (v2 / system
                :relations (v3 / property-ascription
                               :domain v2
                               :range (v4 / faulty)))
    :range (v5 / input)
    :tense present)
\end{verbatim}
\begin{center}
{\small Input for digital circuit example sentence:\\
{\em The faulty system is connected to the input}}\\
\rule{3in}{0.1mm}
\end{center}
\begin{verbatim}
(v1 / sail
    :actor (v2 / ship
               :relations (v3 / property-ascription
                              :domain v2
                              :range (v4 / inoperative)
                              :tense past)
    :destination (sasebo / port)
    :tense present-progressive)
\end{verbatim}
\begin{center}
{\small Input for navy example sentence:\\
{\em The ship which was inoperative is sailing to Sasebo}}
\end{center}
\vspace{-0.3in}
\caption{Input specifications from navy and digital circuit domains\label{navy-circuit-spls}}  
\rule{\textwidth}{0.5mm}
\end{figure*}

Despite the  progress  that  has  been  made  with  the Upper Model as a
potential interface ontology,  it is  still the  case that  the mappings
between grammatical forms and the categories of the Upper Model ontology
are  not  yet  rich  enough  to  ensure  entirely  appropriate  semantic
classifications --- entirely anologously to the case with the explicitly
syntactically  oriented  categories  of   the  {\sc  eurotra-d}   semantic
relations.  In an  attempt to  make the  definitions of  the Upper Model
concepts  more  accessible  to  users   of  the  Penman  system,   these
definitions have been pushed towards an intepretation of the Upper Model
as  {\em predominantly}  a  hierarchy  of  generalizations  about
possible
linguistic realizations in  English.     This  approach permits  a very
straightforward control  of  the  grammar  but  compromises  some of the
semantic integrity. Some  simple examples  of this  may be  seen in  the
following.

In the then current version of the grammar, the 
following clause, which  is  an  example  that arose during development of
Johanna Moore's  {\em  Program  Enhancer  Advisor}  ({\sc pea})
system~\cite{Moore89-thesis}:\footnote{All the PEA examples  were provided in
work by Johanna Moore and Richard Whitney.}
\begin{center}
{\em X is defined as Y}
\end{center}
had to be  constructed  from  a  process   {\em  define}  and  an  
adjunct of
`role-playing' to  produce  the  prepositional  phrase  {\em  as Y}.  This
contrasts with a  more semantically oriented  discrimination of  process
types which could take, perhaps, a process of `defining'  with three
necessary participants, a  {\em definer},  a {\em  defined}, and  a {\em
definition},  and  state  how  these  are  realized  directly.   In  the
realization class view as we have it now, the process of defining has to
be explicitly decomposed semantically  at the level  of the Upper  Model
into a  process  and  a  relationship  of  role-playing.   This  is  not
intuitively obvious: indeed,  a user has  to know  {\em how  the grammar
generates
as-prepositional phrases} in order to arrive at the `correct' Upper
Model classification in order to be able to generate the clause.
This is dangerously close to the amount of low-level syntactic
detail that needs to be provided for a Functional Unification Grammar or
Mumble-86. 

This is not an isolated case. Other problematic assignments in the {\sc
pea} domain include:
\begin{itemize}
\item The process {\em call}, as in ``The boy is called John''.
Presently {\em call} is
classified as a {\em dispositive-material-action} from UM-89, {\em boy}
becomes the {\em actee}, and the name, `John', becomes a {\em
recipient}. No {\em actor} is specified and so a passive construction
appears (due to a then current shortcut defined for the textual
reasoning that
the grammar initiates for selection of active-passive clauses).

\item The  process  {\em  generalize  to},  as  in  ``The  result can be
generalized  to  other  cases''.  Here  {\em  generalize}  is  again   a
straightforward {\em  nondirected-action}  and  {\em  to other cases} is
specified as a {\em  destination} spatio-temporal circumstance  in UM-89
in order to  generate the  preposition.  \end{itemize}  

In all of these cases, the role assignments are only being used in order
to achieve the required syntactic pattern given by the particular  state
of the grammar of the Penman  system: the Nigel grammar of  English.  In
the first example, the model for the  clause being used is that of  {\em
give} since this  class of  verbs is  bitransitive; in  the second,  the
technique adopted is as with the case of {\em define as} above, where  a
circumstantial role is selected purely in order to guarantee the desired
preposition.     Although in  these cases  it is  reasonably clear  that
both grammar and Upper  Model would need  to be extended  to include the
desired process types, in general the {\em theoretical} status of  using
arbitrary assignments of concepts to  the Upper Model and  selections of
roles to  be  expressed  has  not  been  made  sufficiently  clear. This
technique is (or rather  {\em should}) only  be employed when  it is not
possible to extend the grammar and ontology appropriately: only when the
grammar has to be taken as `fixed', e.g., because it is being applied by
a user that  does not  have access  to the  internal organization of the
grammar, is this kind  of strategy defensible.  As a general  technique,
the strategy  has  to  be  strongly  rejected  on  theoretical  grounds.
However, note that  without a  commitment to  {\em semantic}  coherence,
there is little reason not to use  the Upper Model in this way;  we have
already seen the  similar situation  in the  use of  the {\sc eurotra-d}
system of semantic relations where commitment to semantic coherence  has
been explicitly  rejected  in  favor  of more readily operationalizeable
grammatical critera.

Another set of related problems arises when semantically {\em similar}
processes have different syntactic realization. Consider, for
example, the two clause types:
\begin{center}
{\sc x} {\em is like} {\sc y} \\
{\sc x} {\em resembles} {\sc y}
\end{center}
Although a user might wish to place these similarly in the Upper Model,
grammatically they are rather different. The former requires the
grammatical features: \{circumstantial-attribute, manner-participant,
be-intensive\}; the latter has features:
\{circumstantial-ascription, circumstantial-process\}. In the present
Nigel grammar, these are distinguished by the inquiry {\em
circumstantial-ascription-q}, which would need to examine the Upper
Model. Therefore, in order to obtain the differing syntactic structures
a further distinction would need to be set up at the Upper Model
level.

The realization class  view therefore  makes it  difficult for  users to
formulate their  input  specification  to  the  system  unless they know
precisely the form  of linguistic  expression that  they require.  Since
the realizational link between Upper Model categories and Nigel has been
made so tight  for the  very purposes  of achieving  readily describable
criteria, it is sometimes (and increasingly once more users attempt more
varied  modelling)   necessary   to   subordinate   a   concept   in   a
counter-intuitive position simply in order for the language required  to
result.  This certainly undermines the  semantic integrity of the  Upper
Model as  an  interface  ontology  and  moves  the entire classification
towards  a  less  abstract  level   of  information.   It  needs   to be
remembered, however, that only when the grammar is fixed, is a specific,
determinate Upper Model  required ---  furthermore, that  Upper Model is
even partially determined by the particular grammar that is specified.

Finally, one further problem with the interface ontology instantiated by
the  Penman  Upper  Model  lies  precisely  in  the  simplicity  of  the
relationship constructed betweeen domain model and Upper Model. We  have
seen that this is achieved by literally {\em classifying} (in the formal
sense of adding into the  subsumption lattice) domain model  concepts in
terms of the categories from the Upper Model. Following this  operation,
the Upper Model and the domain model form a single inheritance hierarchy
and the domain concepts directly  inherit the possibilities for  surface
realization defined  for  the  Upper  Model  concepts. This operation is
currently performed only  once for  each domain  and, while  simplifying
input expressions, it  means that  the relationship  between domain  and
Upper Model is not  being handled particularly  flexibly. In fact,  once
the classification is complete, the complete ontology can be interpreted
as having collapsed into a mixed ontology of the type described for {\sc
lilog}: both  particular  domain  concepts  and  general  linguistically
motivated concepts  occur  in  the  same  subsumption  lattice.     This
treatment of the relationship  between a potential  conceptual ontology,
containing detail knowledge  of a  domain, and  the interface  ontology,
containing a  semantic  classification  of  possibilities for linguistic
expression, needs to be made considerably more flexible to avoid the
problems of mixed ontologies described both above and below.

\subsubsection{Interface ontologies that are too abstract}

Interface ontologies exhibiting  the problem  of being  too abstract are
more commonly found in small scale  systems:   the problem of not  being
able to specify the  mapping down to  grammar and lexis  in a convenient
and expandable form often prevents large-scale development from  getting
very far. Such projects (e.g.,  {\sc polygloss, acord} and  many others)
begin by  adopting  classes  of  categories  developed,  for example, in
analytical philosophy  or  natural  language  semantics  --- such as the
event types  of~\cite{Vendler67},  temporal  categories  such  as  those
of~\cite{MoensSteedman88},   the    semantico-`conceptual'    predicates
proposed    by~\cite{Jackendoff83,Jackendoff90},    event     structures
of~\cite{Pustejovsky88}, and  many  others.      As  long  as restricted
grammatical  possibilities  are  entertained,  for  example  to   enable
research on  particular  focused  areas  of  semantics-syntax, then such
ontologies are  adequate  ---  even  useful,  since  the focusing allows
greater depth in the semantic account to be achieved. It should also  be
the case, however, that this work then feeds back into more general  and
broader ontology work,  and this  happens much  too rarely.  It is  also
sometimes unclear what the relationship of these ontologies would be  to
a more abstract conceptual ontology --- this may be expressed  formally,
for example, in terms of a  model-semantic theory but the details are
often  left for future work.

\subsubsection{Brief discussion}

It is clear that ontologies of the interface type that are more  closely
bound to language  are nevertheless  most useful  for {\sc  nlp} systems
that want to deal with a wider variety of actual language phenomena. The
increase in abstraction  may not  be so  very great  in comparison to a
desired conceptual ontology, but it is nevertheless better than  working
with grammar and lexis directly. Such  work is also much more  likely to
be stable  in  the  face  of  changing  theoretical  positions  and more
justifiable with respect to actual linguistic data. It is, then, natural
that one further  type of  {\sc nlp}  projects attacking  the problem of
large-scale ontology construction is  that of `dictionary'-oriented
projects,  such   as   {\sc   edr}~\cite{MatsukawaYokota91}   and   {\sc
acquilex}~\cite{Calzolari91}. The {\sc edr} project aims at producing  a
`concept dictionary' containing  400,000 `word  senses' for  English and
Japanese, and {\sc  acquilex} is  concerned with  producing a  re-usable
`lexical knowledge base' that classifies entries according to taxonomies
of semantic  categories  and  relations  between  those categories. Both
projects have constructed sizeable semantic taxonomies relying  strongly
on differences  in  lexico-grammatical  realization  for  the categories
adopted.   The  taxonomy  organization  and  categories  found  in  {\sc
acquilex} have similarities  to the  view of  lexical semantics proposed
by~\cite{Pustejovsky91-cl} where, again, {\em oppositions} in linguistic
behavior are an  essential motivating  criterion. Another  large project
partly leading up to this work, and  now related to the {\sc kbmt}  work
mentioned  above,   was   the   MIT   Lexicon  Project  where  extensive
classification of lexemes was undertaken  on the basis of  the differing
grammatical   patterns    that    the    lexemes    may    enter
into.
% (e.g.,~\cite{Levin89,Levin-fc}).

Although the  construction  of  large  knowledge  bases at this level of
abstraction is bound to offer a  definite improvement in our ability  to
rely  on  linguistic  motivations  in  future  ontology  design,   their
availability will not of  itself bring about  that design.  It  is still
necessary to consider methodologies for  using such information so  that
appropriate ontologies for  general {\sc  nlp} use  can be  constructed.
Therefore, in the  next section  I will  relate the  kinds of ontologies
that we have  seen in  this section  to compatible  linguistic theories.
Without a broader view of what  is being done {\em linguistically}  when
categories for a particular kind of ontology are proposed, I believe  it
is unlikely  that  progress  will  be  made.   As long as the categories
developed are sufficiently close to  the surface details of  language to
remain objectively verifiable, i.e., remain in the realms of syntax  and
lexico-grammatically    oriented     interface     ontologies,    useful
classifications can be constructed.  For more abstract levels,  however,
the  support  of  theory  become  crucial  for  defining  methodologies,
questions, and possible solutions.

\section{Linguistic support (or otherwise) for the ontology types}
\label{ling-theories}

In this  section,  I  will  follow  the  ordering  of  the discussion of
ontology types of the previous section: i.e., first linguistic  theories
compatible with  the  design  of  mixed  ontologies  will  be mentioned,
followed by the  kind of  linguistic theory  that is  more supportive of
distinct interface and  conceptual ontologies.     I will  not raise the
issue here  of  the  relationship  between  `conceptual  ontologies' and
possible linguistic theories, since one of the defining phrases that  is
often  used  about  this   level  of  abstraction   is  its  very   {\em
extra-linguistic}ness. This does,  however, depend  on the  view of  the
linguistic system that  is adopted  and I  will mention  something about
this later.   Finally in  this section, I discuss  some
disadvantages of the former approaches when considered as a  methodology
for developing the kinds of resources necessary for {\sc nlp} systems.

Before beginning the  discussion, I  should however briefly  note the
motivation
for  an  exclusion  of  forms  of  semantics  such  as  situation
semantics, model-theoretic semantics of  various
kinds, etc. below.   Such accounts  are not immediately relevant  to the
discussion at hand precisely because  they have not been  concerned with
the construction of representations that are directly supportive of {\em
ontologies}. That  is,  regardless  of  whether  the  formal  account of
semantics  proposed  in  some  particular  framework  contains  sets  of
predicates  that  are  of  mixed  ontological  status,  or  are   purely
conceptual, or  purely  (linguistically)  semantic,  we find one crucial
component of ontological engineering  missing.     Those categories  are
not typically  built  up  into  subsumption  lattices  of  sorts sharing
various general properties of use for further domain classification.  It
is clear that many of these theoretical approaches could easily move  in
this direction, and with the increased use of sorts in linguistic theory
at all levels  of description  some first  steps have  been taken (e.g.,
~\cite[p78]{SagPollard91},~\cite{Nerbonne91}). However, as pointed out
by~\cite{OnyshvekychNirenburg91}:
\begin{quote} \small
`The crucial point is that in order to have an explanatory power, the
atoms of [a] meaning representation language must be interpreted in
terms of an independently motivated model of the world. Moreover, if any
realistic experiments have to be performed with such an {\sc nlp}
system, this world model (sometimes called an {\em ontology}) must be
actually built, not only defined algebraically.'
\end{quote}
Therefore, until the
problem of ontology construction on a realistic scale itself
becomes an issue for an
account, that account remains  of less central concern for the current
discussion. 

\subsection{Mixed ontologies and linguistic theory}

The closest linguistic approaches to support mixed  ontology
design such as that found   in {\sc  lilog} are,  perhaps
surprisingly, those
compatible   with    the    work    of~\cite{Jackendoff83,Jackendoff90}.
Jackendoff adopts the position that the semantic level of representation
with which he  is concerned  is also  {\em conceptual},  i.e., common to
modalities such  as  language  and  vision~\cite{Jackendoff83}.       As
pointed out  by~\cite{Herweg91},  approaches  that  directly link syntax
with conceptual interpretation now occupy a rather standard position  in
mainstream linguistics and so  there are many  approaches that could  be
described. That of Jackendoff is probably one of the most developed  and
well known in this direction, although there are also similarities to be
drawn with work in Cognitive Linguistics~\cite{Langacker87,Talmy87}  and
directions such as that of~\cite{Wierzbicka88}. All of these  approaches
share an orientation to language  as an instrument for  revealing facets
of conceptual organization. This is stated most clearly by Jackendoff in
terms of what he terms the {\em Grammatical Constraint}:
\begin{quote} \small
`\ldots it would be perverse not to take as a working assumption that
language is a relatively efficient and accurate encoding of the
information it conveys.  To give up this assumption is to refuse to
look for systematicity in the relationship between syntax and
semantics. A theory's deviations from efficient encoding must be
rigorously justified, for what appears to be an irregular
relationship between syntax and semantics may turn out merely to be a
bad theory of one or the other.'~\cite[p404]{Jackendoff83}
\end{quote}
Given his equation of semantic structure and conceptual structure, this
becomes largely equivalent to statements such as the following
describing the basic claim of of cognitive linguistics:
\begin{quote} \small
``... across the spectrum of languages, the grammatical elements that
are encountered, taken together, specify a crucial set of concepts.
This set is highly restricted: only certain concepts appear in it, and
not others\ldots [This] set of grammatically specified notions
collectively constitutes the fundamental conceptual structuring system
of language. That is, this cross-linguistically select set of
grammatically specified concepts provides the basic schematic
framework for conceptual organization within the cognitive domain of
language.'' \cite[p165/6]{Talmy87}
\end{quote}
This position also appears in the approach of  Pustejovsky to the relation between
lexemes and their interpretation in context; as he writes,
\begin{quote} \small
`The meaning of  words
should somehow reflect the deeper,  conceptual structures in the  system
and the domain it operates in.   This is tantamount to stating that  the
semantics of  natural  language  should  be  the  image of nonlinguistic
conceptual principles (whatever their structure).'~\cite[p410]{Pustejovsky91-cl}
\end{quote}
These approaches are all described by the first architecture depicted in
the diagram of Figure~\ref{3-ontologies}. Each suggests that there is  a
portion  of  the  conceptual  ontology  that  has  a  direct  linguistic
connection  and  that that portion  should  have  just  the  same  kind   of
organization as  the  rest  of  the  conceptual  ontology. A
specification of the semantics of some expression is simultaneously a
(possibly partial) specification of a conceptual specification. Again,
this state of affairs receives a very explicit description from
Jackendoff: 
\begin{quote} \small
`This account of the syntax-semantics correspondence gives a principled
account of the level of ``argument structure'' found in various
versions of GB and LFG ... - a level of linguistic representation
that lists the arguments of a verb, with or without their 
$\theta$-roles. Such a list can now be simply constructed from the set of
indices in the conceptual structure of the verb, and there is one
index per syntactically expressed argument... In short, ``argument
structure'' can be thought of as an abbreviation for the part of
conceptual structure that is ``visible'' to the syntax.'~\cite[p404/5]{Jackendoff83} 
\end{quote}

By virtue of the Grammatical Constraint, therefore, Jackendoff adopts  a
very  close  binding  of  linguistic  analysis  and  categories  at  his
semantico-conceptual  level  of  representation:   available  linguistic
realizations  and  patternings   lead  directly   to  the   positing  of
corresponding   categories   and   relationships   at   the   level   of
semantic/conceptual structure. In Jackendoff's case, the linguistic evidence admitted is
organized in terms of $ \overline{\rm X} $-theory
\cite{Chomsky80,Jackendoff77}
and so  close correspondences appear between categories of this theory
and categories of the semantic/conceptual structure. In particular, he
states that:
\begin{enumerate}
\item ``... every major phrasal constituent in the syntax of a
sentence corresponds to a conceptual constituent that belongs to one
of the major ontological categories.''
\item ``... the lexical head X of a major phrasal constituent
corresponds to a function in conceptual structure --- a chunk of the
inner code with zero or more argument places that must be filled in
order to form a complete conceptual constituent. The argument places
are filled by the readings of the major phrasal constituents strictly
subcategorized by X.''~\cite[p67]{Jackendoff83}
\end{enumerate}
Thus, he suggests the following approximation to conceptual structure
for the sentence {\em The man put the book on the table} \cite[p68]{Jackendoff83}.

{\small \[ \left[ \begin{array}{l}
            \sc event \\
            \sc put(  \left[ \begin{array}{l}
                             \sc thing \\ \sc the \;\;  man
                          \end{array} \right] ,

                  \left[ \begin{array}{l}
                             \sc thing \\ \sc the \;\; book
                          \end{array} \right]\\ ,
                 
                  \left[ \begin{array}{l}
                            \sc place \\
                            \sc on ( \left[ \begin{array}{l}
                                           \sc thing \\ \sc the \;\; table
                                        \end{array} 
                                  \right] ) 
                         \end{array} \right] )
          \end{array} \right] \]}

This  structure,  if  we  ignore  the  textual  information  represented
abbreviated here with  {\sc the},  shows striking  similarities with the
input specification described earlier for  Penman (cf.
Figures~\ref{mumble-comp} and~\ref{navy-circuit-spls}). The structure may be
glossed as stating that a predicate {\em put} of type {\em event}  holds
over three arguments: the first two are of type {\em thing}, the  latter
is an {\em on}-relation of type {\em place}. Each of the predicates  are
taken  to  be  defined  as  semantico-conceptual  categories   motivated
primarily by linguistic patterning.  Further examples of  the motivation
of semantico-conceptual  categories  from  linguistic  evidence  is  the
following list of example categories offered by Jackendoff:
\begin{tabbing} \small
\=a.n\=How did oyu cook the eggs?nn\= \kill
\>   \> \ \ \ \ \em Interrogative probe \> \em supports category:\\[2mm]
\>a. \> What did you buy? \> [{\sc thing}] \\
\>b. \> Where is my coat? \> [{\sc place}] \\
\>c. \> Where did they go? \> [{\sc direction}] \\
\>d. \> What did you do?   \> [{\sc action}] \\
\>e. \> What happened next? \> [{\sc event}] \\
\>f. \> How did you cook the eggs? \> [{\sc manner}] \\
\>g. \> How long was the fish? \> [{\sc amount}] 
\end{tabbing}
Subsequently, further
categories of differentiations are made working from intuitions on the
meanings of sentences and their
constituents supported by example sentences. Moreoever, 
analogously to the perceived relationship between syntactic
structures and rules for their well-formedness, Jackendoff takes the
position that the inter-relationships between the semantic/conceptual
categories will also be expressed in terms of well-formedness rules.
An example for the category [{\sc path}] is as follows
\cite[p166]{Jackendoff83}: 

{\small \[ {\sc [path]} \rightarrow \left[ Path 
                      \left\{ \begin{array}{l}
                                \sc to \\ \sc from \\ \sc toward \\ 
                                \mbox{\sc away-from} \\
				\sc via 
                              \end{array} \right\}
                       \left( \left\{
                              \begin{array}{l}
                                [_{Thing} \; {\rm y}] \\ 
                                {\rm [}_{Place} \; {\rm y}]
                              \end{array}
                       \right\} \right)
                      \right]
\]}

The combination of a number  of rules such as  these begins to define  a
hierarchy  of  interrelated   categories  analogous   to  the   standard
hierarchical organization that I  have assumed appropriate  for ontology
construction. 

A comparison of Jackendoff's  semantico-conceptual categories with,  for
example,  the  superficially  very  different  categories  arising  from
cognitive linguistics  is  very  illuminating  concerning  the role that
motivations from  language  can  play  for  ontology  construction.  The
general  methodology  of  proponents  of  cognitive  linguistics  is  to
examine `grammatical' elements --- however these come to be defined  ---
in order to  uncover the  conceptual organization  they presuppose.  For
example, Talmy offers the  following break down  of the {\em  this/that}
distinction in English.

\begin{quote} \small
`A closed-class element of this type specifies the location of an
indicated object as being, in effect, on the speaker-side or the
non-speaker-side of a conceptual partition drawn through space (or
time or other qualitative dimension).'~\cite[p168]{Talmy87}
\end{quote}
This is summarized as:
\begin{itemize}
\item a `partition' that divides a space into `regions'/`sides'
\item the `locatedness' (a particular relation) of a `point' (or
object idealizable as a point) `within' a region
\item (a side that is the) `same as' or `different from'
\item a `currently indicated' object and a `currently communicating'
entity.
\end{itemize}
By sampling across a wide range of languages the Cognitive Grammarian
compiles a list of such distinctions and attempts to provide internal
organization and structure rooted in a presumed linguistically relevant
area of conceptual organization. The flavor of this organization can be
seen in the following examples of proposed categories from Talmy.
\begin{description} \small
\item[Dimension] ``The category of `dimension' has two principal
member notions, `space' and `time'. The kind of entity that exists in
space is --- in respectively continuous or discrete form --- `matter'
or `objects'. The kind of entity existing in time is, correspondingly,
`action' or `events'\ldots ''~\cite[p174]{Talmy87}. This is schematized as:
\begin{center}
\begin{tabular}{||l||l|l||} \hline
\em dimension & \em continuous & \em discrete \\ \hline
space : & matter & objects \\
time :  & action & events  \\ \hline
\end{tabular}
\end{center}

\item[Plexity] `Plexity' is a generalization of notions such as
singular and plural to cover actions also. For example:
\begin{tabbing}
nnn\=a.nn\=multiplexnnn\=A bird flew in.nn\=He sighed (once). \kill
\>   \>         \> \em matter              \> \em action \\
\> a. \> uniplex \> \em A \rm bird flew in. \> He sighed \em (once). \\
\> b. \> multiplex \> Bird\em s \rm flew in. \> He \em kept \rm sigh\em ing.
\end{tabbing}

\item[Boundedness] `Boundedness' is a generalization of notions 
such as mass and count with respect to nouns to include again actions
in addition to objects. This Talmy relates to {\em imperfective} and
{\em perfective} and similar terms in the treatment of verbs.
Essentially, ``[w]hen a quantity is specified as `unbounded', it is
conceived as continuing on indefinitely with no necessary
characteristic of finiteness intrinsic to it. When a quantity is
specified as `bounded', it is conceived to be demarcated as an
individual unit entity.'' (\cite[p178]{Talmy87}). Similar, far more formal,
expressions of this idea can now be found in a number of approaches
(e.g.~\cite{Krifka89}). 

\item[Dividedness] ``A quantity is `discrete' (or `particulate') if it
is conceptualized as having breaks, or interruptions, through its
composition. Otherwise, the quantity is conceptualized as
`continuous'.''\linebreak \cite[p180]{Talmy87} 
\end{description}
These  categories hold  of  a given `quantity'   simultaneously and so
classify that quantity   along the  dimensions  described. Moreover,
different
linguistic consequences are intended to  follow from each distinction.
Although there are many interesting distinctions
suggested which could help enrich proposed ontologies along a number of
dimensions,  the lack of an accepted, detailed
grammatical framework nevertheless limits the generalizations that can
be found.  Langacker claims that:
\begin{quote} \small
``\ldots basic grammatical categories such as {\bf noun}, {\bf verb},
{\bf adjective}, and {\bf adverb} are semantically definable. The
entities referred to as nouns, verbs, etc. are symbolic units, each with
a semantic and a phonological pole, but it is the former that determines
the categorization. All members of a given class share fundamental
semantic properties, and their semantic poles thus instantiate a single
abstract schema subject to reasonably explicit characterization. A noun,
for example, is a symbolic structure whose semantic pole instantiates
the schema {\sc [thing]}\ldots In a similar fashion, a verb is said to
designate a {\bf process}, whereas adjectives and adverbs designate
different kinds of {\bf atemporal relations}.'' \cite[p189]{Langacker87}
\end{quote}

Although with the proposed conceptual categories restricted in this  way
to follow from grammatical categories that are so directly `observable',
i.e., often inflectional  and word-based  such as  singular and  plural,
mass  and  count,  nouns  and  verbs,  etc.,  one  would  not  expect  a
particularly  rich  ontology,  in  fact,   a  large  number  of   finely
differentiated categories are set up --- primarily on the  basis
of contrastive examples  that do  {\em not}  rely on  detailed syntactic
analysis. This shows  conclusively the  value of  examining a  very wide
range of natural occuring examples,  in contrast to the  oft criticised
(e.g.,~\cite{Rohrer86}), but nevertheless  still prominent,  tendency in
mainstream linguistics  to  study  constructed  examples  in  areas that
illuminate the currently fashionable linguistic phenomena. Nevertheless,
the lack  of  a  formally  specifiable  mapping  between  the categories
proposed  and  linguistic  realization   renders  the  consequences   of
establishing any  particular  set  of  categories  almost  impossible to
investigate and this  is certainly  less of  a problem  in a contrasting
account such as  that of  Jackendoff where  the relation  to a  detailed
account of grammar and lexis is  always clear. The value noted  above of
being able  to  {\em  test  out}  and  justify  proposed  categories for
ontologies  formally  applies   here  strongly.   Jackendoff  is   able,
therefore, even on  the basis  of rather  limited linguistic  breadth of
motivation,  to  suggest   a  more   detailed  set   of  categories  and
interrelationships.    The   semantico-conceptual   representations  are
substantially more abstract than syntactic classes (as evidenced by  the
generalizations that they permit to be drawn) but are nevertheless  tied
reasonably precisely with possibilities for linguistic expression.    An
ideal situation would therefore  be to have  a very broad,  detailed and
formally specified  grammar,  capable  of  describing  very fine-grained
grammatical and lexical differences.

Even despite the lack of formally specified mappings to linguistic  form
within  cognitive  linguistics,  there  has  still  been  at  least  one
significant application of  its proposed  concepts in  a computationally
context. This  is  in  their  use  to  provide a system of {\em semantic
features} for  stating  meanings  to  be  preserved  across languages in
machine    translation     \cite{Zelinsky-Wibbelt87,Zelinsky-Wibbelt88}.
Although the work suffers from the lack of explicit definition that
the conceptual categories have so  far received --- making  it difficult
for coders using the semantic features reliably to classify the meanings
that are involved --- this situation may be improved significantly by  some
current   work    in    progress\footnote{For    example   by   Cornelia
Zelinsky-Wibbelt and  Wiebke  Ramm  of  IAI/Eurotra-D on syntactic tests
that coders could apply to  resolve difficult cases.} which  is intended
to improve the necessary connection between the semantic categories  and
their linguistic  realization.    The  situation  applying  Jackendoff's
categories in a computational  context has, as  would be expected,  been
more straightforward. A number of proposals  have been made for such  an
application,  and  some   have  been   implemented.       For   example,
\cite{Meteer88}  comments  on  the  possible  organization  of  abstract
linguistic terms at the  text message level  for the sentence  generator
Mumble-86 that a system such as  Jackendoff's could provide and we  have
already seen that  both~\cite{Dorr87,Dorr90-thesis}'s work  on the  {\sc
unitran}    translation    system    and    approaches    within    {\sc
kbmt}~\cite{NirenburgLevin91}   have   implemented   aspects   of    the
semantico-conceptual structure.\footnote{Further analogous areas of research
which often fall somewhere between the explicit grammatical foundation
attempted by Jackendoff and the, until now, more impressionistic
linguistic motivations of Langacker and others, include the large body
of work on the `conceptualization' and linguistic expression of
spatial-temporal information, e.g.:~\cite{Herskovits86,BierwischLang89}
and many others.}

\subsection{Interface ontologies and linguistic theory}
\label{ling-motivation}

In contrast to the accounts of the previous section, the separation of
information found in the interface ontology and a more
abstract conceptual ontology  is consonant  with theoretical positions
that assume a higher degree of {\em stratification} in the linguistic
system.
The   mixed    ontology    view    goes    well    with    a    standard
syntax-semantics-pragmatics distinction where  `semantics' includes  the
conceptual representation  and  `pragmatics'  provides  procedures  that
operate over the semantico-conceptual  representation to produce  active
interpretations in context. In this  sense, pragmatics is not  a further
stratum in a linguistic system and has a distinct theoretical status  to
that of  syntax  or  semantics.   In  contrast  to  this,  the interface
ontology architecture  suggests  at  least  a  three-way  stratification
between lexico-grammatical  information,  semantic  information,  and  a
contextualizing level of `conceptual' information. Each of these  strata
appears  to  have  rather  similar  formal  properties:    most  of  the
information of each, for example, would appear to be representable as  a
subsumption  lattice  defined  over   sorts,  possibly  augmented   with
structured  inheritance.\footnote{Current   work   in  information-based
syntax makes this point for syntax~\cite{PollardSag87}.}

I have already mentioned  one view of  the linguistic system  that seems
compatible with this  stratification:    the  approach to  semantics and
context proposed by~\cite{Bierwisch83,BierwischLang89} that acted as one
influence  for  the  {\sc  lilog}  design  ---  even  though  the  final
specification of the ontology within {\sc  lilog} does not seem to  have
remained in the spirit of this  theory.  Within the linguistic model  of
Bierwisch, conceptual representations  are maintained  strictly separate
from semantic  representations,  and  semantic  expressions  are used to
constrain construction of conceptual expressions during  interpretation.
Thus, `words'  (actually  lexicogrammatical  patterns)  are  related  to
semantic forms  which  determine  functions  from contexts to conceptual
structures. The distinction between the two levels in this kind of {\bf
two-level semantics} is nicely summarized by Michael Herweg as follows:
\begin{quote} \small
`Semantic representations are structured configurations of semantic
units which, on the one hand, are determined by the grammatical system
of the language in question and, on the other hand, are grounded in ---
or motivated by --- the conceptual system. \ldots Conceptual
representations are structured configurations of conceptual units, which
are mental representations of certain aspects of the external
world.'~\cite[p152/3]{Herweg91}
\end{quote}
The two classes of categories --- the semantic and the conceptual ---
thus have very different theoretical statuses and allow very different
kinds of motivations. This is therefore precisely the kind of
structuring of the linguistic system that one requires to support the
use of interface and conceptual type ontologies. 

The  most   successful   of   the   interface  ontologies  described  in
Section~\ref{ontology-examples}, the Penman Upper  Model, clearly has  a
natural relation to the stratification found in this kind of  `two-level
semantics'.  For example, the sorts of the Upper Model are determined by
the grammatical system (concretely, the  Nigel grammar component of  the
Penman system) as is required.  Although there is also a relationship to
be drawn  with  accounts  that  are  explicitly  seeking  {\em semantic}
organizations  closely   linked   to   language   regardless   of  these
organizations'    further    embedding     at    higher     levels    of
abstraction,\footnote{Or, alternatively, by seeking  an embedding in  an
account such as model-theoretic semantics to `bottom out' in a  formally
specifiable way.}, the relation between a proper view of the Upper Model
and two-level semantics becomes even  closer when we examine  instead of
the Upper  Model,  rather  the  {\em  theoretical position} of which the
Upper  Model  is  only  a  very  partial  instantiation:  i.e.,  that of
systemic-functional
theory~\cite{Halliday61,Halliday78,MatthiessenHalliday89-sfg}.

Systemic-functional theory  is  a  highly  stratified general linguistic
theory with  respect  to  which  the  Penman  text  generation  and  its
descendents have  been,  and  continue  to  be,  developed. In some ways
perhaps analogously to the situation in  {\sc lilog}, many aspects of  the
current  implementation   of   the   Penman   system  are  not  accurate
instantiations of that theory. Of particular importance here is the very
instantiation of  the  concept  of  linguistic  strata  --- since it is
precisely this construct which is  necessary for motivating the  kind of
multi-levelled representation that we  find in interface  ontologies and
their contextualizing conceptual ontologies. 

\begin{figure*}
\rule{\textwidth}{0.2mm}
\epsfig{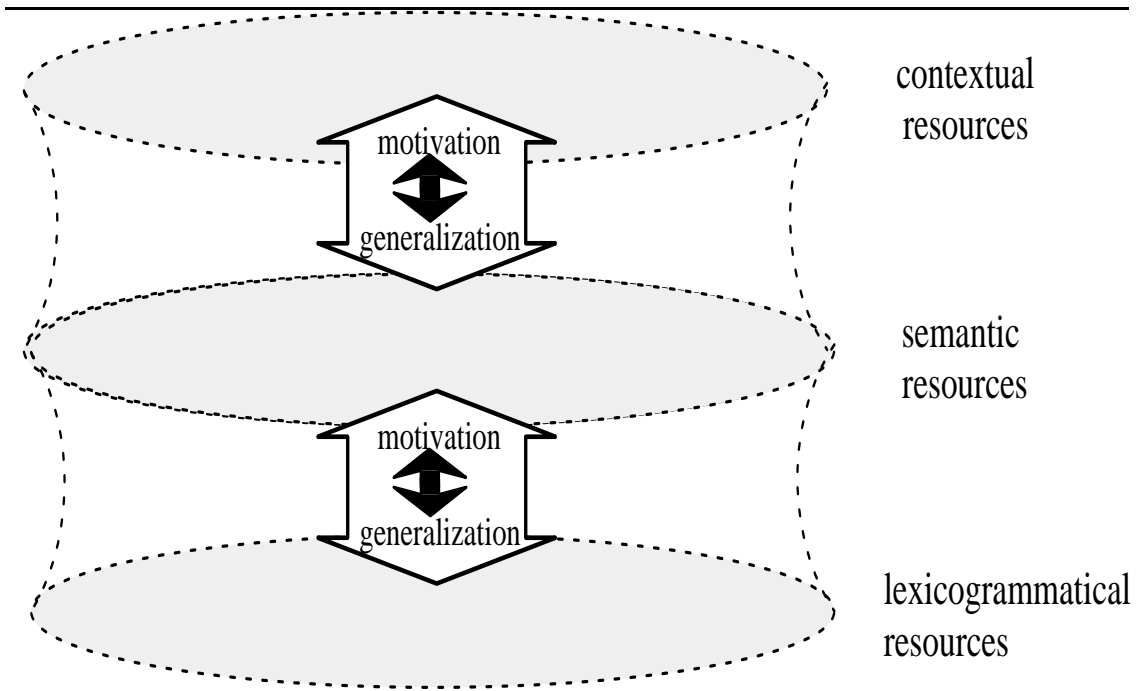}{90}{150}
\caption{Stratification with systemic-functional linguistics}
\label{strata}
\rule{\textwidth}{0.2mm}
\end{figure*}

The notion of stratification  in systemic-functional theory  is depicted
diagrammatically  in  Figure~\ref{strata}.   The  linguistic   system is
broken  down  here  into  three  strata:  lexicogrammar,  semantics, and
context.  Between each  stratum the  same relationship  --- that of {\em
realization} ---  holds.   Systemic-functional  theory  is essentially a
{\em functional} theory, i.e., one that is concerned crucially with  the
functions that  language  fulfills  in  particular  contexts,  and  this
informs the understanding of the realization relationship between strata
as follows.  Each higher-level (i.e., more abstract) stratum is seen  as
providing the  {\em  functional  motivation}  for  the  next lower-level
stratum; and each lower-level  stratum is seen  as providing a  resource
that {\em  generalizes}  across  the  possibilities  of  the next-higher
stratum~\cite{Halliday78}.  This gives  us a  more detailed  view on how
strata in  the  linguistic  system  interact  than that usually found in
stratified accounts.  Additionally, each higher-level stratum is seen as
{\em contextualizing} the levels beneath.

The organization of  the Penman-style  architecture version  of systemic
theory instantiates the stratification as follows.  Nearest the  surface
there are {\em realization  statements} of syntagmatic  organization, or
syntactic form.   These  statements  are  classified  in  terms of their
potential for  expressing  communicative  functions  that  are  realized
grammatically, such  as  asserting/questioning/ordering, active/passive,
etc.: this denotes paradigmatic organization and is represented in terms
of a {\em grammatical system  network}.  This organization captures  the
possible alternatives that  are available  given any  choices that  have
already been made;  i.e., a  collection of  `paradigms' of  the form  `a
linguistic unit of type A is either an A of functional subtype X, or  an
A of functional subtype Y, \ldots, or an A of functional subtype Z'  are
given.   At  each  level  these  subtypes  are  disjoint  and  serve  to
successively  classify   linguistic   units   along   ever  more  finely
discriminated dimensions.  This formulation of classifications in  terms
of  increasingly   fine   discrimination   is   in   systemic-functional
linguistics termed  the  principle  of  {\em delicacy}.  The grammatical
communicative functions are  then in  turn {\em  motivated} by  semantic
distinctions that  classify  semantic  circumstances  according  to  the
grammatical features which are appropriate to express those  situations:
this classification is the combined responsibility of {\em choosers} and
{\em inquiries}~\cite{Mann83-anatomy}.  Finally,  the possibilities  for
classification that  the  inquiries  have  are  defined  in terms of the
abstract   ontology   of    the   Upper    Model.    In    relation   to
Figure~\ref{strata}, then,  the  Penman-style  architecture represents a
computational instantiation only for the {\em lower} two strata and  the
relationship between them.

\begin{figure*}
\rule{\textwidth}{0.2mm}
\epsfig{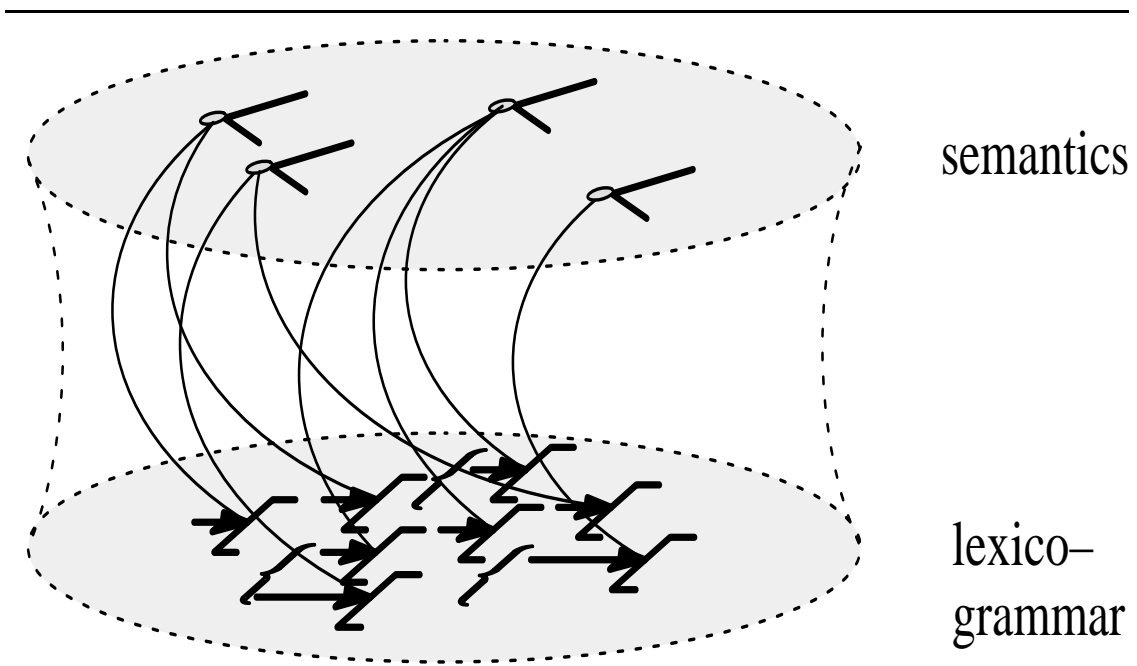}{90}{150}
\caption{Covering the grammar semantically}
\label{covering}
\rule{\textwidth}{0.2mm}
\end{figure*}

While at a rather general level  very similar to the breakdown  proposed
by Bierwisch, the systemic-functional account also goes into more detail
about the  internal  organization  of  each  stratum. It is this feature
which is largely responsible both for the more abstract status that  has
been achieved  for  the  sorts  of  the  Upper  Model  and for the early
adoption of the  principle of  motivating sources  on the  basis of  the
grammar.  Not only  is all  grammatical variation  captured by  abstract
choices between  minimal  grammatical  alternatives,  but  also all such
abstract choices must have explicit motivations, or semantic conditions,
defined.  Only  then  is  the  grammar  fully  defined as a resource for
grammatical  expression:    we  have  to  know  what  each   grammatical
possibility is an expression  of. This has  naturally given rise  to the
notion of {\em covering  the grammar} in  terms of a  set of motivations
for each choice that the grammar offers. This is depicted graphically in
Figure~\ref{covering}. The  categories  necessary  for this motivational
covering are then organized into sorts in a subsumption lattice --- thus
defining the Upper Model.

It is worth  noting that  this provides  a very  strong methodology  for
interface ontology  construction.      Until  a  grammar  alternation is
explicitly connected into a  motivational relationship, the  alternation
is considered to  be only  formally (in  the sense  of {\em  linguistic}
form) defined. The grammar in fact acts as a (highly structured) list of
phenomena that  require  semantic  motivation.   In  addition,  the {\em
functional} organization of the grammar  itself goes a long  way towards
providing a useful pre-classification of syntactic phenomena so as to be
amenable to  systematic  semantic  interpretation.  The  extra  boost in
abstraction that the  grammar offers  is responsible  for the  increased
level of abstraction that the Upper Model has already achieved. 

I have noted,  however, that  the Upper  Model does  not instantiate the
full organization required by the  theory.  Some of the  consequences of
this have  already  been  mentioned.   For  example,  the  {\em upwards}
relations of the  Upper Model  to context  has not  been modelled within
Penman in  terms  of  a  `realization'  relationship:  domain models are
directly subordinated to the  Upper Model hierarchy.   I will return  to
this and some other problems  below.  Importantly, since the  full input
of the theory has not yet been  taken into account, we have available  a
number of possible  directions for  development that  may provide  a far
more  sophisticated  implementation  of  both  interface  and conceptual
ontologies.

\subsection{A comparative evaluation}

In this  section,  I  will  apply  the  possible  linguistic theoretical
underpinnings that could be provided for the different ontology types in
order to consider those ontology  types more critically. I  will suggest
here that there are clear reasons for dispreferring accounts that  adopt
a mixed ontology  approach. Subsequently,  in the  next section,  I will
discuss  possible  future  developments  for  interface  and  conceptual
ontologies drawing further on the connections to theory established.

We have seen that the type of account based on a style of  argumentation
such as  that  of  Jacken\-doff  manages  to gain abstractness while still
maintaining contact with details of linguistic realization. I have noted
that the increase in abstraction  is a generally necessary  property for
improving the functionality of {\sc  nlp} systems. One of  the principle
differences  between  such  an  account  and  the  linguistic   theories
supportive of interface ontologies was in the degree and explicitness of
stratification. One  can  ask  the  question,  therefore,  is  there any
evidence for the more stratified view  of the linguistic system?  If  it
proves necessary, or  beneficial, to  differentiate between  information
that is particularly linguistic  and the kind  of information sought  in
accounts of  real-world,  commonsense  knowledge,  then a mixed ontology
will not be  sensitive to  this. A  very important  issue to address is,
therefore, whether the  selection between  a mixed  ontology and  a more
differientiated set of  interrelated ontologies  is one  which is  still
open to debate, or arbitrary ---  or are there grounds for  deciding for
one architecture over the other.

\subsubsection{Populating a mixed  ontology}

If we  assume  that  we  have  an  account  such  as  that  proposed  by
Jackendoff, possibly augmented  by a  range of  concepts from  cognitive
linguistics  with  a  more   formally  expressed  relationship   to  the
lexicogrammar, it is still the case that there the resulting ontology is
not yet very  large.  The  number of  general sorts  that occur  in, for
example, ~\cite{Jackendoff90} (i.e., not  the conceptual equivalents  of
lexical items, which  seem to  be introduced  freely), is  less than 40:
these include predicates such  as {\sc event,  state, be, orient,  path,
go, with,  from,  to,  towards,  inch}oative,  {\sc react, aff}ect, etc.
Most conceptual items are decomposed into these `primitives'. I will not
discuss whether or not these  items are good candidates  psychologically
for conceptual primitives, but relying  on this small set  of categories
is unlikely  to  capture  many  generalizations of linguistic expression
when a broad lexicogrammar is  considered.     If we include  experience
such as that obtained within the development of the {\sc lilog} ontology
or  the  Upper  Model,  many  intermediate  sorts  will  express  useful
generalizations over distinct linguistic patterns. Relying on a  smaller
than necessary set of categories either misses generalizations or places
more  work  on  the   mapping  with  lexicogrammatical   form.       The
methodological question  arises  of  how  the  sort  hierarchy  is to be
extended beyond  the  very  general  categories  that  most  attempts at
ontology construction assume as basic on intuitive grounds.

The primary source of  evidence for extension  is the classification  of
lexicogrammatical    patterns.
%     such     as     that    carried    by,
%e.g.,~\cite{Levin89,Levin-fc}.    
This  posits  semantic  features  that
co-occur with particular classes  of Lexical Conceptual  Structures. But
these classes are constructed on  the purely {\em syntactic}  linguistic
behaviour of the investigated lexemes.       This, while being the  best
methodology available and  one I  have defended  throughout this  paper,
cannot itself  be  expected  to  give  rise to {\em conceptual} classes.
Only  the  assumption  that  such  semantic  patterns  are  simultaneously
conceptual makes this plausible:  there  is no obvious connection  to be
drawn between aspects of domain and commonsense knowledge and  lexically
derived categories. The latter are often subject to criticism for  being
too shallow  even  for  an  interface  ontology:   they must appear very
unlikely candidates for a conceptual ontology. As we have found with the
problems with  the  Upper  Model  (Section~\ref{um-intro}),  there is no
guarantee  that  particular  domain-motivated  categories  will   choose
lexically-motivated categories that belong to a consistent more  general
ontological type.  More often items belonging to very different  lexical
classes are treated as semantically equivalent for speakers'  expressive
purposes.    Representing this in a  single ontology then requires  that
concepts  may  be  consistently  classified  along  the  two  dimensions
simultaneously: which complicates the formal properties of the resulting
ontology considerably since exactly what may be inherited where  becomes
unclear.

This is  shown  concretely  in  the  linguistically motivated evaluation
that~\cite{Lang91} undertakes  for  {\sc  lilog}.  There he examines the
sorts proposed for the  ontology according to  the kinds of  motivations
accepted for  their  inclusion.  He  finds  the following differentially
motivated  sorts  all  combined  in  the  single  subsumption   lattice:
\begin{itemize} \item `Conceptually based  sorts' which are  included on
extra-linguistic (conceptual) grounds.

\item  `Text  base  specific  sorts'  which  are
concepts corresponding  to  special  vocabulary  items  required  by the
particular domain  and  text  with  which  {\sc  lilog} as a project was
concerned.  

\item `Sorts projected from  the grammar' which are  notions
found in  the  grammar,  such  as  {\em preposition}, transferred to the
ontology.  

\item `Sorts of mixed  origin' which are concepts  where both
extra-linguistic and  linguistic  criteria  are involved.  
\end{itemize}
This mixing of motivations organizes itself loosely according to
the  vertical  and  horizontal   dimensions  in  the   hierarchy.  Thus,
\begin{quote} \small  `The  vertical  structure  of  the sort hierarchy,
which is  based  on  the  subsumption  relation,  draws  mainly  on  the
availability of  corresponding  linguistic  labels  categorized as nouns
\ldots or as verbs \ldots However, the horizontal dimension of the  sort
hierarchy, that is the selection of subsorts to be assigned to a  common
supersort, is  mainly  determined  by  features  that  emerge  from  our
extra-linguistic conceptual knowledge  of objects  and spatio-temporally
specifiable   events    or    situations   \ldots'~\cite[p466/7]{Lang91}
\end{quote}

Lang  shows  the   following  problems  for   the  resulting
organization in a single subsumption lattice that this inconsistency,  or
variety, of motivations for concepts in the hierarchy creates.    First,
since  extra-linguistic  or  conceptual  criteria  are  less  than  well
understood, there is  a degree  of arbitrariness  in the categorizations
that appear.   Second,  it is  never clear  from the  concepts that  are
found in the hierarchy alone whether they  are to be expected to have  a
corresponding linguistic effect or  not.     Third, the  co-existence of
distinct  kinds  of   concepts  means   that  the   precise  meaning  of
`subsumption' with  respect  to  particular  cases is underspecified ---
different kinds  of  concepts  have  different  relations  between their
`wholes' and their  `parts' and  until this  is clarified  it is unclear
what kind  of  subsumption  actually  holds.   These differences entail
different formal  properties  so  that  different  objects  can call for
different inheritance properties.      Thus,  for example, a  supposedly
general `part-whole' relation is intended  sometimes as `is a  component
of', sometimes as `is spatially included in', somtimes as `pertains to',
sometimes  as   `inalienable   possession',   sometimes   as  `alienable
possession', etc.    This  range of  possibilities makes  the inferences
that in  fact  follow  from  any  statement  in  the  ontology  far more
difficult to  foresee  and  substantially  complicates  in  any case any
axioms for inference that are designed. 

This can also be seen concretely
in many versions of semantics where  a mixed ontology is relied  upon ---
in order to handle the very flexibility of the relationship between  the
concepts that are to function  for the linguistic expression,  and those
which  are  not,   complex  and   often  unconstrained   mechanisms  are
introduced: the  `projective  inheritance'  of~\cite{Pustejovsky91-cl}  and
many    instances     of     `type     coercion'     as     used     by,
e.g.,~\cite{SagPollard91,Pustejovsky91-cl} are probably  prime examples  of
this, but there are many others. A mixed ontology is, therefore, a very
weakly constraining theoretical construct, which does not provide
optimal assistance either for theory construction or for system
construction.

\subsubsection{Stratification}

The  just  mentioned  flexibility  of  relationship  between  conceptual
categories and the categories that are determinative of their linguistic
realization is  a  very  typical  property  of  a  relationship  between
linguistic strata. It is this very flexibility, in fact, which  provides
the primary linguistic  evidence for  stratification.  As  an example of
this, consider the following  issue of ontological  design.   

Regardless of whether a mixed ontology  is adopted or not, some  portion
of some ontology is assumed which  offers an expression of the  chunking
that language  expects  and  demands  of  knowledge  if  it  going to be
expressible  through  the  grammatical  and  lexical  resources  of  the
linguistic system.  One question that  can be asked, therefore,  is:  is
the information in a conceptual ontology that will support this chunking
{\em already organized in  this way or  not?} If it  is then it  will be
straightforward to  construct  a  mechanism  such  as  that suggested by
Jackendoff above, whereby one simply  `takes a view' on  some conceptual
structure  and   already   has   a   specification   of   the   semantic
predicate-argument structure which can in  turn control the grammar  and
lexis to  produce  appropriate  results.  If  not, however, then some
reorganization of the structure will be necessary. In all examples
that are presented of alleged conceptual structures that are already
appropriate for direct lexicogrammatical realization, e.g., by viewing
as predicate-argument structure, we can make the following observations:

First, the lexical items and class of grammatical patterns
appropriate is
already so highly constrained as to follow directly  from the
expression; for example,
\begin{center} \small
[$_{\mbox{\footnotesize \rm State}}$ {\sc orient} ([$_{\mbox{\footnotesize \rm Thing}}$
{\sc weathervane}], [$_{\mbox{\footnotesize \rm Path}}$ {\sc north}])]
\end{center}
as one reading for the sentence: {\em The weathervane pointed
north}~\cite[p74]{Jackendoff90}. Certain variability in
lexicogrammatical expression will be produced by the mapping rules of
syntax formation, but other decisions, including: the choice of word for
the concept {\em weathervane} given that the hearer might not know what
a weathervane is, or that the sentence may be uttered among
world-experts on the subject of weathervanes who would normally select a
far more restrictive description, etc. have already been built into the
description. Widely differing selections of possible expression
according to text type, register, formality, situation, time
availability (cf.~\cite{Hovy88-book,BatemanParis89}) are excluded.

Second, the  {\em granularity}  of the  corresponding language has
also been  built  into  the  description.   For  example, we know that a
sentence is  going  to  be  produced  (or  if the linking rules are good
enough: a sentence or a  nominalization) rather than a  short discussion
of the wind's effect  on an object  whose position of  equilibrium under
the pressure  of  the  wind  serves  as  an  indication  of  the  wind's
direction. A nice example\footnote{Due to Karin Haenelt.} of maximal
flexibility here might be the
difference, for example, in the language produced in response to the
conceptual real-world category {\em beer} for the purposes of a
dictionary entry, e.g.,
\begin{quote} \small
`{\bf Beer} is a bitter alcoholic drink made from grain. There are a lot
of different kinds of beer.' [Collins {\sc cobuild} English Language
Dictionary, 1987]
\end{quote}
and that produced for the purpose for an entry in an
industrial chemical encyclopedia, which goes on for 40 pages. 

The response to both of  these problems within the  semantico-conceptual
approach is straightforward: the differences are expressed beforehand in
the semantico-conceptual  organization  and  are  produced by conceptual
processes for information organization  and management. But  this misses
the generalization  that  {\em  regardless}  of  the  information  to be
expressed that  same {\em linguistic} granularity is imposed: there  will
be a set of descriptions of  some predicate with an argument  structure,
including specifications  of  participants  and  circumstances.  The two
sentences concerning  beer  in  the  dictionary  and the hundreds in the
encyclopedia all exhibit  the same  kind of  organization.  Knowledge is
variable      scale,      but       language      is       predominantly
fixed-grain,\footnote{Apart from  the  resources  for combining clauses,
nominal groups, etc.  into  `complexes', which are  not relevant to  the
current argument.} as defined by the  grammar.  This means that for  all
the knowledge available in the semantico-conceptual ontology, there need
to be  construction  mechanisms  available  which  convert some selected
fragment of the information,  of any scale,  and produce an  appropriate
sized chunk of semantico-conceptual structure for motivating a sentence.

With unconstrained inferencing  across the  knowledge base  this may  be
achievable by inheriting constraints back from the grammar and  checking
the equivalence of constructed semantico-conceptual structures with  the
originally selected fragment.    But,  crucially, for all  such selected
fragments, the same class of `semantico-conceptual' paraphrases will  be
potentially available:      i.e.,  those licensed  by their  grammatical
expressability.  Furthermore, also regardless of the originally selected
semantico-conceptual fragment, the lexico-grammatically licensed set  of
`semantico-conceptual'  specifications   govern   specifiable   sets  of
inferences that operate {\em only on such} specifications:  for  example
the  inferences  that   determine  the   textual  variations   that  are
appropriate        when        realizing        the        specification
lexicogrammatically~\cite{BatemanMatthiessen93},  that  certain
abstract semantic classifications apply for which there is no conceptual
evidence~\cite{Schriefers90}, and others.   Thus, not representing  this
distinguished  set   separately   fails   to   capture   a   significant
generalization about the organization of the linguistic system as a
whole.\footnote{It is also engenders dubious {\sc nlp} system design;
factoring out the commonalities in a separate stratum is analogous to
the following application of object-oriented programming:
\begin{quote} 
``In an object-oriented application \ldots the system uses predefined
mappings from objects to the routines that know how to process those
objects (or can choose among different routines depending on the
context). The efficiency of using predefined mappings for known types
comes in drastically reducing or entirely eliminating search; the onus
is put on the developer to define the decisions available to a type at
each level, rather than presenting all options at all times and letting
a search procedure find the best one.'' \cite[p6]{Meteer89}
\end{quote}
This is also one property of using an interface ontology such as
Meteer's or the Upper Model.} 

Finally, it is worth emphasizing that this flexibility between strata is
typical and not unique to the relation between semantics and  conceptual
levels of representation.   The  relationship between, for  example, the
Upper Model and the lexicogrammar already exhibits much of the same kind
of flexibility. For example, the expressive resources of the grammar  of
nominal groups is not restricted to the single grain-size of sorts  that
are subtypes of an Upper Model sort {\em object}. It is equally possible
to realize  Upper  Model  classified  {\em  events} as nominal groups or
configurations of events as single clauses if the textual conditions are
appropriate.  \cite{BatemanParis89}  present  other  examples  of   this
theoretical flexibility for other categories  in the grammar. It  is not
at all surprising given the theoretical similarity, therefore,  to find
exactly this kind of flexibility again  between the sort lattice of  the
semantic ontology and that of the conceptual ontology.

\begin{figure*}
\rule{\textwidth}{0.2mm}
\epsfig{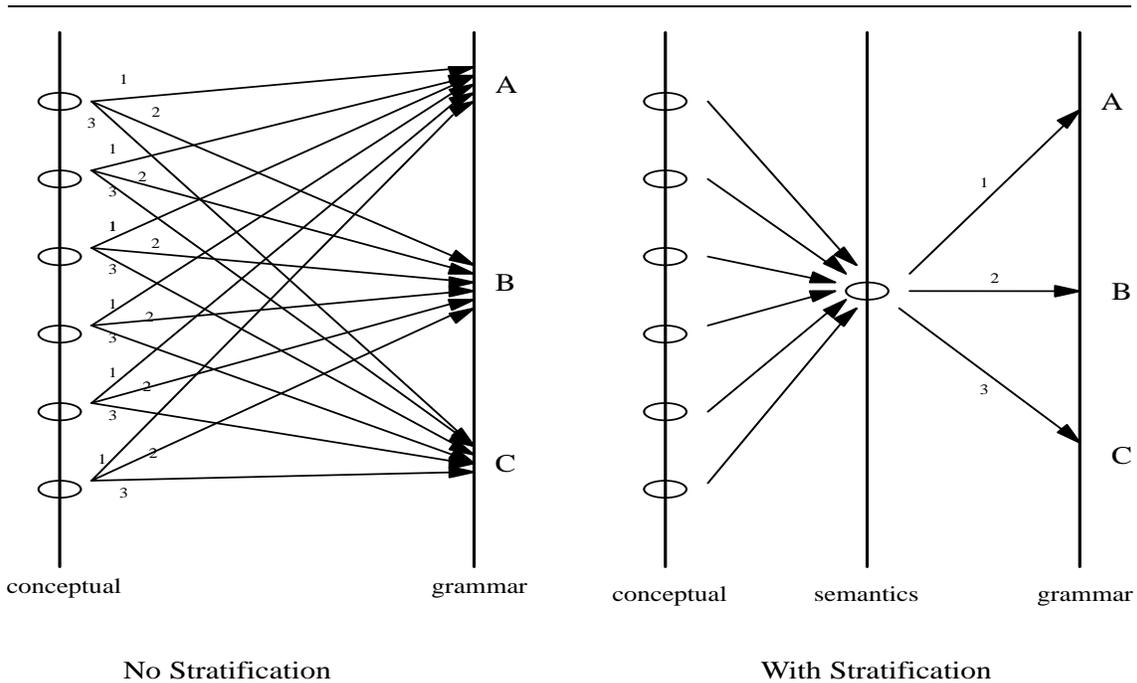}{90}{150}
\caption{Capturing generalizations via stratification} 
\label{stratification}
\rule{\textwidth}{0.2mm}
\end{figure*}

This    discussion     of     stratification     is     summarized    in
Figure~\ref{stratification}.  Here we see three strata and the  repeated
variability in expression that any selected {\em semantic} specification
has.  Crucially, the  common, reoccuring  coding possibilities  that are
available for all elements from the conceptual stratum are not  repeated
at that level, but are factored into a single statement at the level  of
the semantic  interface  with  a  mapping  from  sorts at the conceptual
stratum to  sorts  at  the  interface  stratum.    Not representing this
generalization both guarantees a complication of the theory and makes  a
usable {\sc nlp} system based on the theory unlikely.  Again, the  power
of the theory to bring methodological and contentful constraints to bear
on system design is compromised.

\section{Some Principles and Methods; and some former puzzles resolved}

The discussion up  to this  point has  attacked mixed  ontologies on the
basis that  they  are  internally  inconsistent,   and  has criticised
the
non-statificational linguistic accounts underlying such mixed ontologies
on the  basis  that  they  fail  to  capture theoretically important and
practically useful generalizations. Both weaknesses have one consequence
in common:   they  provide a  seriously reduced  set of  constraints for
ontology design and construction. Since one purpose of this paper is  to
suggest principles and  methodologies for  ontology construction,  mixed
ontologies and under-stratified  linguistic accounts  are clearly  to be
avoided. The kinds of ontologies most appropriate for
{\sc nlp} systems
and for which  linguistic support  needs to  be sought can now be
restricted to the following two types.
\begin{itemize}

\item Type $O_i$: an abstract  semantic organization underlying our  use
of grammar and lexis that  is motivated on essentially  {\em linguistic}
grounds and that acts as  a complex interface between  lexicogrammatical
resources and  higher-level  strata  in  the  linguistic  system --- the
categories of this  interface should  be maximally  general, i.e., apply
across distinct real-world situations, but specific enough to  maximally
constrain possible lexicogrammatical expression.

\item Type  $O_c$:   an  abstract  organization  of real-world knowledge
(commonsense or otherwise)  that relates  downwards to  the interface to
lexicogrammar. 
\end{itemize}

With these restrictions in place, I will now go further and suggest some
particular guidelines for ontology construction.   In order to do  this,
however, it is also  necessary to make  some further commitments  to the
kinds of information that will be made available at particular levels in
the linguistic system. The reason for  this is that the linguistic,  and
particularly lexico-grammatical,  constructs  are  essential for guiding
ontology design. This follows the increasingly wide range of  linguistic
theories that are returning  to the position  that the relation  between
grammar and semantics is not arbitrary;  we saw a selection of  these in
Section~\ref{ling-theories}                                       above,
e.g.,~\cite{Langacker87,Talmy87,Wierzbicka88,Jackendoff83,Halliday78}.
If we accept this, then it is also to be accepted that the selection  of
{\em particular  accounts  of  lexicogrammar}  has  consequences for the
subsequent ontology design. Since  such consequences cannot  be avoided,
it makes sense to make selection decisions in ways which will  maximally
help in the task of ontology construction overall. I will distinguish
between decisions in the following two areas:
\begin{itemize}
\item type of grammar
\item contents of grammar
\end{itemize}
and will make some firm suggestions for the former and discuss
the consequences of differences that arise in the latter.

First, we can note that the most successful interface ontology developed
so far is probably that of the Upper Model. The Upper Model has achieved
both a detailed account and a generally applicable account. We can  ask,
then, what is it about its underlying theoretical organization that  let
this occur? Second, we can further note that although the Upper Model is
the most detailed instantiation  of an ontology  of type $O_i$  that has
been developed, is nevertheless not  a full instantiation of  the theory
on which it  is based.  It is  therefore worthwhile  considering briefly
what additional constraints the theory could bring to bear if it were to
be more fully implemented. 

The kind of  grammar on  which the  abstractions proposed  in the Penman
Upper Model is easily classified. It is a {\em  paradigmatic-functional}
grammar  exhibiting   the   standard   Hallidayan  {\em  metafunctional}
diversification~\cite{IFG,Matthiessen90-lexcartog,MatthiessenBateman91}.
This means that  it is  organized, firstly  around {\em  choice} --- the
paradigms  of  grammatical  constructions   that  stand  in   functional
opposition  ---  and  second  around  a  factorization  of  that  choice
according to its semantic  motivation:    is the  choice to do  with the
propositional content of  the linguistic  entity to  be classified ({\em
ideational}), is  the  choice  to  do  with  the  textual placing of the
linguistic entity to be classified ({\em textual}), or is the choice  to
be classified  as  to  do  with  the  interpersonal relationship between
speaker and  hearer  or  with  the  attitude  of the speaker towards the
information expressed by  the linguistic  entity ({\em  interpersonal}).
The motivations for the choices provides hypotheses concerning the sorts
necessary for  controlling  those  choices.   The  Upper  Model has been
derived by considering  motivations for  those choices  {\em exclusively
assigned to  the  ideational  metafunction}:   there  is  no  mixing  of
categories  across  metafunctional   domains.\footnote{Although  it   is
perfectly possible  to  imagine  applying  the  same `grammar-as-filter'
methodology on underlying motivational ontologies as carried out for the
ideational metafunction ---  cf.~\cite{Bateman90-catalina} for  examples
of     this      applied      to      the      textual      metafunction
and~\cite{MatthiessenBateman91} for general discussion --- the resulting
organizations of information have very different properties.} 

This builds into the design  of an ontology motivated  in such a way  as
the Upper Model the following features.

First, we require an ontology that is significantly more {\em  abstract}
than syntactic realization  classes. I  have already  suggested how this
has been achieved with the  Upper Model.   The grammar,  being organized
in terms  of  a  functional  classification  of  possible constraints on
constituency structure, is {\em already} more abstract than constituency
structure {\em  per  se}.   Further  classification across the paradigms
uncovered is  then  automatically  more  abstract  and  achieves  a
generalization  across  particular   lexico-grammatical  contexts   that
supports a greater flexibility of  expression of input expressions.  The
strict relationship to the grammatical stratum also makes sure that  the
kinds of mixed sorts that~\cite{Lang91} finds and criticises in the {\sc
lilog}  hierarchy  {\em  cannot}  occur:  either  an  (interface,  i.e.,
semantic)  ontological  category   has  a   specified  consequence   for
lexico-grammatical expression or it is not accepted.

Second, given the stratification suggested by the theory the Upper Model
is automatically only the `next level  up' in the linguistic system:  it
is an ontology  strongly connected  to grammar  below.  It  does not, by
itself, provide the necessary  organization of higher  level ontologies.
Thus, in short,  we see  that an  organization closely  reminiscent of a
two-level semantics  is  automatically  achieved,  and  that  {\em both}
levels require ontologies. 

Third, we have  seen that  it is  a design  goal that  an ontology be as
general as  possible  ---  that  it  helps  with  classification  across
domains, tasks,  and  applications,  but  also  be substantial enough to
provide a  rich  scaffolding  for  domain  description.  This raises the
question: How can we guarantee that a proposed ontology is as general as
we require?  We  can now  see that  ontologies such  as the Upper Model,
which are based  on motivations  from grammar,  are {\em  guaranteed} to
have  the  domain-independence  required  of  them.   Since  ontological
categories are motivated  by the  grammatical distinctions  (and not  by
more arbitrary  lexical  collections  found  in  a  given domain), those
categories are forced to be {\em at least} as general as the grammatical
categories.     It     is,     therefore,     very     unlikely     that
\cite[p462]{Klose-vonLuck91}'s claim that the {\sc lilog} ontology has a
`more domain-independent status'  than the  early version  of the  Upper
Model  described  in~\cite{Mann-etal85-janus}  would  apply  to  current
versions of the Upper Model.

\begin{figure*}
\rule{\textwidth}{0.2mm}
\epsfig{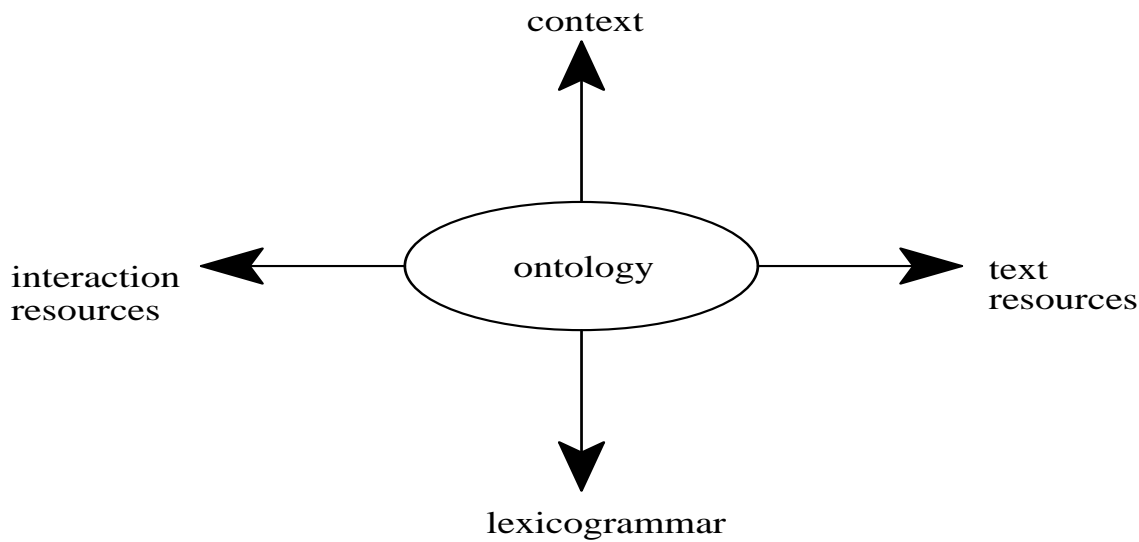}{90}{150}
\caption{Capturing generalizations via metafunctionality} 
\label{4way-generalizations}
\rule{\textwidth}{0.2mm}
\end{figure*}

But we can go further and move beyond the kind of generalizeability that
refers simply  to  domain-independence  ---  which  is generalizeability
`upwards' in the linguistic  system, and beyond  generalizeablity across
the lexicogrammar  ---  which  is  generalizeability  `downwards' in the
linguistic system.     When  we also  consider the  {\em metafunctional}
organization of  the  linguistic  system  posited by systemic-functional
theory,  then  we  can  see  that  generalizations  both  across   `text
instances' and across `speech functions'  are also guaranteed ---  i.e.,
generalizations `horizontally' across the same stratum of the linguistic
system.                        This   is    depicted   graphically    in
Figure~\ref{4way-generalizations}. These  constraints  rule  out certain
other potential sorts from the ontology, e.g., sorts concerned with  the
particular appearance of an entity at a given position in a text or with
the speaker's attitude  towards an  event.  Certain  of the  sorts found
in~\cite{Meteer89}'s interface ontology are good examples of the  former
kind. Having such sorts requires reclassification of domain  information
whenever a domain object is used  in a text, since the  textual statuses
of domain objects changes over the development of a text --- i.e.,  from
new to given, from theme to rheme,  etc. This change of course needs  to
be represented: the point is  that representing such information  in the
interface ontology again {\em mixes} very different kinds of information
--- although this time on a `horizontal' dimension across the linguistic
system rather than a `vertical' one.

The kind  of  grammar  that  we  employed  as the initial motivation for
guiding the development of the Upper  Model has, therefore, gone a  long
way towards ensuring that the  properties desired of ontologies  obtain.
But an area of flexibility in the description then arises from the  {\em
depth} of grammatical description, i.e., the {\em contents}, rather than
the type.  Particularly within the systemic-functional approach, lexical
descriptions  are  seen  as   more  specific  versions   of  grammatical
descriptions --- there is no difference in {\em kind}.  Thus, if we push
lexicogrammatical description  further  in  the  direction  of lexis, we
automatically push  further  the  depth  of motivating semantic ontology
constructs that are needed.   This bifurcation in  potential description
needs more theoretical work before we can make any firm statements about
whether it is more helpful to pursue one at the expense of the other, or
whether they should be pursued in parallel as has been the case with the
more general area of the grammar.

We can now also consider some possible improvements and explanations for
some  awkward   phenomena/intuitions   that   have  previously  hindered
ontological engineering. For  example, if  there is  a stratification of
the kind argued for, why is it that suggestions for conceptual structure
that have been put forward in a  number of approaches appear also to  be
candidates for representation as sorts in the interface ontology? ---
When the categories of the Upper Model, for example, are examined, many
similar classes to the proposed `conceptual' ontology work are to be
found. 

To give a concrete example of this, ~\cite[p474]{Lang91}, after  careful
discussion concerning the  problems of  a mixed  ontology, defines  some
basic assumptions concerning  the structure  of the  conceptual ontology
drawn from earlier work, including~\cite{BierwischLang89}. With  respect
to these assumptions, he outlines  the following set of  conceptual {\bf
domains} which are to form basic subsorts of the conceptual ontology:
\begin{tabbing} \small
\hspace*{0.1in} \= D$_4$: time intervals; \= D$_5$: events;
\hspace*{0.23in} \= D$_6$: attitudes \kill
\> D$_1$: objects; \> D$_2$: substances; \>
D$_3$: locations; \\
\> D$_4$: time intervals; \> D$_5$: events; \> D$_6$: attitudes
\end{tabbing}
We can also note here similarities with some of the classes above
from~\cite{Jackendoff83,Jackendoff90,Langacker87,Talmy87}, etc.
But these are also sorts already found, for example, in the Penman Upper
Model, where they have been entered purely on the grounds that they are
necessary to directly constrain possible grammatical realizations. Is it
the case that the claim we saw above  by~\cite[p133]{Gust91}  that:
`there are
continuous variations between semantic forms and conceptual structures'
is, after all, true? Can we introduce strict stratification and still
account for the intuition that these concepts indeed function at
different strata?

Lang already suggests that there  may be certain genuinely  `linguistic'
features that function definitionally  for features at  the `conceptual'
stratum:
\begin{quote} \small
`\ldots the representation of nouns like {\em Ofen} [oven], {\em
Fahrzeug} [vehicle], {\em Boot} [boat] in the lexicon contains a
specific component {\sc purpose} (hence, an element of our linguistic
knowledge) by means of which the sort {\em Nutzgegenstand} [article for
practical use] in the knowledge base is being accessed. This is but one
example of how linguistic aspects of lexical representation can be made
use of in defining ontological sorts in the knowledge
base.'~\cite[p470]{Lang91} 
\end{quote}
Other `genuine linguistic features' that Lang suggests for the basis of
the ontological distinctions include: `bounded object' vs. `non-bounded
object', `concrete object' vs. `abstract object' --- both very similar
to other theoretical accounts. We can now go further and {\em explain}
the relation between the linguistic (semantic) ontology types and the
conceptual ontology types as follows.

All of the  reasoning that  we have  applied to  the development  of the
Upper Model ontology with respect to its motivation in the lexicogrammar
can be  applied  precisely  to  the  relation  between  the  Upper Model
ontology and some higher stratum ontology. This follows as a consequence
from the theoretical statement of  the nature of realization  within the
stratified linguistic account.   This  means that we  will need to  find
motivations for the semantic interface  ontology sorts.  It also  means,
however, that we can make use of the realization relation starting  from
the standpoint of the  higher stratum and interpret the  status of
the semantic interface ontology  as {\em generalizing}  across different
conceptual stratum situations; cf.  Figure~\ref{strata}. Thus, for  both
the lexicogrammar with respect to  the semantic interface ontology,  and
for the  semantic  interface  ontology  with  respect  to the conceptual
ontology,  it  is  likely  that  the  {\em  more  general  intra-stratal
organization of the lower stratum is likely to be echoed in the  overall
intra-organization of the higher stratum}.   This gives us the  observed
link between  constructs  that  are  motivateable  as  general  semantic
concepts  and  constructs  that   appear  to  organize   the  conceptual
hierarchy. There is, then,  no `mixing' of  the categories of  different
strata, there is just a resonance  or echo of categories at  one stratum
taken up at another. 

Given both this theoretical and practical binding of the contents of the
different strata, is it clear why there is then a certain {\em  tension}
between  strata  ---  as~\cite[p462]{Klose-vonLuck91}  note  from  their
experience with  the  {\sc  lilog}  ontology:  
\begin{quote} \small `The
tension between linguistic  and inferential  demands on  the modeling is
alive and  forces  compromises  on  both  sides.'     
\end{quote} 
I have suggested that the  kind of  view of  realization between  strata found  in
systemic-functional linguistics, where  there is  both a  practical {\em
and} a theoretical `pulling' in  both directions --- upwards  to context
and downwards to (experiential) lexicogrammar, offers an appropriate way
of  operating  within  this  tension  between  strata.    The  resulting
methodology then uses the tension to {\em help constrain}  organization
decisions for the construction of  interface ontologies that are  useful
for {\sc nlp} and to remove the need for genuine `compromises' where  an
inappropriate category is postulated at one level because that level  is
insufficiently functionally differentiated from others.

It is clear, however, that we know a great deal more about possibilities
for ontologies of type $O_i$ than we do about ontologies of type  $O_c$.
Moreover, given the results  of the last  section, perhaps we  know even
less than we thought --- clearly conceptual categories are now sometimes
best reappropriated to a more abstract semantic type.       This is a
less
than ideal situation --- particularly  given the view of  stratification
shown in  Figure~\ref{strata}  and  the  established  dialectic  between
strata.  Because  the  realization  relationship  between strata is {\em
bi-directional}, we should be able  to use a higher-strata  to constrain
our accounts at a lower-strata.  But  the fact that we know  very little
about the  higher-strata  in  this  case  removes one source of possible
constraint.

\begin{figure*}
\rule{\textwidth}{0.2mm}
\epsfig{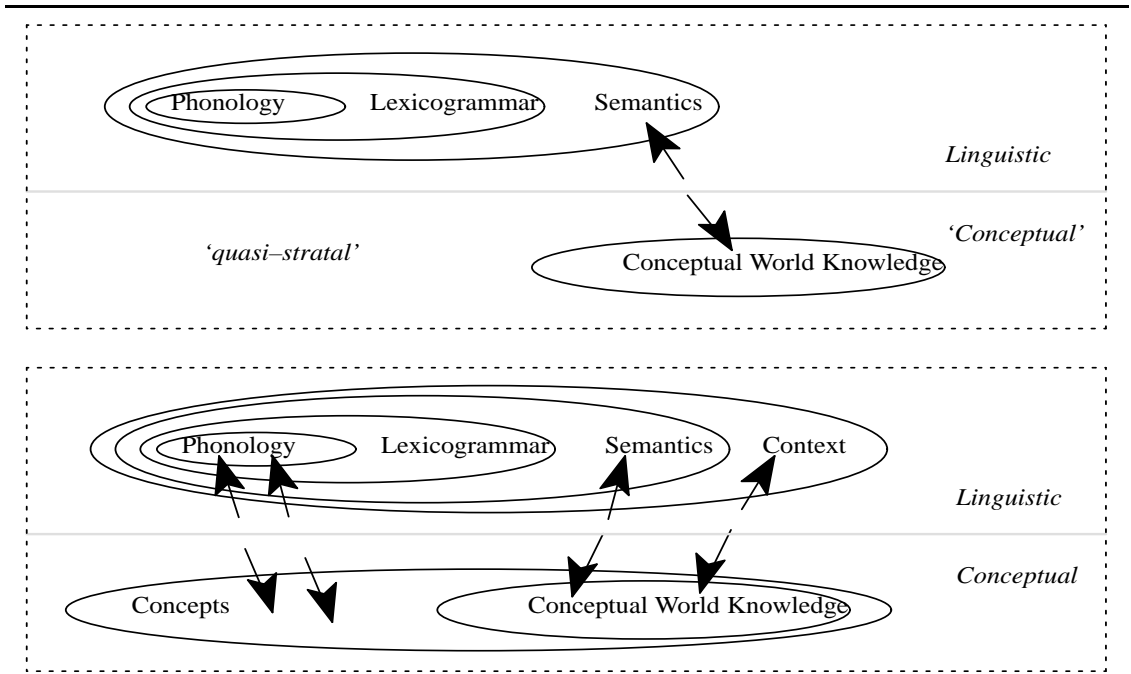}{90}{150}

\caption{The relationship between semiotic descriptions and conceptual representation\label{no-concepts}}
\rule{\textwidth}{0.2mm}
\end{figure*}

Finally, here,  however,  I  will  draw  attention  to  one  interesting
consequence for the {\em status} of the higher-stratum ontology when  we
take  into   account   the   bi-directionality   of   the  inter-stratal
relationship. Since there  is no  difference assumed  in the theoretical
status of the levels related by the interstratal relationship, one might
ask how it  is that  the interface  ontology is  termed `linguistic' and
`semantic', whereas the higher-stratum ontology is `non-linguistic'  and
`conceptual'. I  believe  that  a  far  more  appropriate  view  of  the
relationship is  as  depicted  graphically in Figure~\ref{no-concepts}.
All strata  that  stand  in  an  interstratal  relationship  of the kind
explored and used in this paper should be seen as {\em semiotic}  levels
of greater and lesser degrees  of abstraction.  The conceptual  ontology
thus becomes more  of a  {\em contextual}  ontology, with  context being
interpreted in the sense of a  level of social situation ---  closely in
line with,  for  example,~\cite{Halliday78}.      There  is,  then,  the
additional question of how this entire complex of inter-related levels of
semiotic descriptions  relates  to  the  supporting conceptual system of
human psychology. This  is probably  a very  different kind  of relation
than realization --- although  it will probably  again turn out  to be a
dialectic relationship rather than a one-way determination. This puts us
in the position to  criticise some of  the conceptual sorts  proposed by
Lang  on  exactly  the  same  grounds  that  he  has  criticised   mixed
ontologies. For example, alongside the  above mentioned domains, all  of
which  may  be  more  plausibly  ground  in  the   conceptual/perceptual
system,~\cite[p474]{Lang91}   places:      `social   institutions    (law,
administration,  marriage,  education)'   and  communicative   behaviour
(etiquette, conversation, group dynamics). Such a mixed set of sorts  is
unlikely to  form  a  very  stable  or  usable  ontology: it is probably
crucial to begin  to refine  further our  levels of  ontology, and their
interactions, so that the mistakes made at the least abstract levels  of
ontological engineering are not just  repeated again, at the  next level
`up'. More detailed statements must, however, be left to future research!

\section{Summary, conclusion and final words}

This discussion of this paper has considered the  notion of `ontology'.
Starting from
the view that an ontology is an organization of the world --- which  has
been  approached   by   `naive   physics',   `conceptual  dependencies',
`commonsense (meta)physics', and others --- I drew attention to the fact
that such accounts  do not  bring strong  methodological and substantive
constraints to  bear  on  ontology  construction.    Also unclear is the
relationship of such ontologies to language.  The gap is often so  large
that this  level  is  too  abstract  to  have any direct relationship to
required forms of  expression. Contrariwise,  this gap  also leads  to a
weakening of the  discriminative power  of the  constraints that  can be
brought to  bear  by  linguistic  patternings.   Concretely,  then,  one
cannot, for example, generate natural language directly from such levels
of description  without  resolving,  or  `fixing'  an  immense number of
degrees of freedom that remain unaddressed (often quite rightly, if this
is seen as a  conceptual ontology) in  the ontology itself.  Much of the
work that an {\sc nlp} system requires to be done is, therefore,  simply
not taken into consideration by the abstract ontology.  Such  ontologies
are also, because of their abstractness, difficult to populate  reliably
--- if  sizeable  and  potentially  distributed resource construction is
undertaken, as it  increasingly is,  then this virtually  guarantees poor
intercoder consistency.    In  short,  such  ontologies  are of very
limited value  for {\sc nlp} work.

These problems  have  been  noted  by  some  of  those  who  have sought
principles for ontology design (cf.~\cite{SkuceMonarch90}) and those who
need real shareable resources (as for example in machine translation ---
cf.~\cite{SteinerReuther89}). The only solution  that has been  found to
this endeavor is to place more reliance on {\em language} as a source of
constraint.    For  this  reason,  then,  views  on  language  and   the
organization of the linguistic system become crucial for ontology design
that is appropriate for {\sc nlp}. Moreover, only by taking views on  the
linguistic system that are  maximally supportive of  the functionalities
required of ontologies  can we  avoid problems  of lack  of abstractness
(i.e., being  dominated  by  linguistic  form)  and problems of too much
abstractness (i.e., being dominated  by semantic theories  of particular
areas that lack connection to linguistic form). In short, ontological
engineering faces the following dilemma: interface ontologies
\begin{itemize}
\item need to be abstract, large-scale, re-usable information
classification devices, 
\item but they cannot be too abstract,
\item or too near syntax,
\item and need to be constrained from language.
\end{itemize}
The theoretical assumptions and resulting organizational decisions  that
I have pursued in  this paper appear  to offer a  very practical way  of
preceding within this state of affairs.  I have also shown that  several
other beneficial properties for {\sc nlp} systems are derivable from the
abstract organization of the linguistic system that  systemic-functional
theory posits.

The paper has presented  for broader debate  a round of  discussion that
begun in  the  context  of  the  developing  ontology of the Penman text
generation system.    This  work, beginning  with the  pre-computational
descriptive account, called  the {\em  Bloomington Lattice}  by Halliday
and  Matthiessen has  passed  through  several  instantiations  in
computational form.  Now  future work  will have again  consider  bringing
together the linguistic descriptive account --- reworked to a new  level
of detail in~\cite{HallidayMatthiessen-BL}  --- and  the computational
model. It is to  be hoped that  this approach will  build on the  former
success of the  Upper Model,  simultaneously moving  us in  some of  the
directions that I raised as responses to problems with the Upper  Model.
Thus, I have not suggested that the Upper Model we find in Penman is the
`general solution' to  ontological engineering  --- there  are many more
criticisms to  be  made  of  this  ontology, again mostly concerning the
extent to  which  it  succeeds  as  an  instantiation of the theoretical
principles that underlie it.  The {\em function} of the ontology is also
more finely  circumscribed  than  many  others  ---  but  again strictly
according to the underlying theory. We are  not yet at a stage where  an
ontology can be accepted,  even pragmatically for  the needs of  current
{\sc nlp}  systems,  as  `complete':      what  is  more at issue is the
development of  appropriate  methodologies  for constructing ontologies,
and here  again  constraints  offered  by  the  linguistic system are of
paramount  importance.         The   linguistic   system,   when  viewed
appropriately, gives  a  rich  multidimensional  set  of  constraints on
adequate and  appropriate  designs  for  computational  systems.     The
principle dimensions applied in  this paper were  those of {\em  strata}
and {\em metafunctions}.  This by  no means exhausts the  possible input
of  the  theory,  however.    For  further  dimensions  of  the  theory,
see~\cite{MatthiessenBateman91}; for additional  examples of  using
these   dimensions   to    constrain   computational    system   design,
see~\cite{Bateman-etal-ip}. I hope that the paper has suggested some  of
the benefits of employing such linguistic motivations, and that  further
attempts to apply wider sets of motivations will help us in the future.

\section*{Acknowledgments}

This paper is based on  the work of the  natural language group at  ISI,
including over the  years the  input of  Bill Mann,  Ed Hovy,  Christian
Matthiessen, Bob Kasper, Johanna Moore, C\'ecile Paris, Richard Whitney,
and Robert Albano.           Further  theoretical discussion with  Erich
Steiner,  J\"org  Sch\"utz,  and   Cornelia  Zelinsky-Wibbelt  (IAI   --
Saarbr\"ucken), Elisabeth  Maier,  Elke  Teich,  Renate Henschel and Leo
Wanner (IPSI), Martin Emele and R\'emi Zajac (IMS -- Stuttgart) and with
participants  at  the  Technical  University  of  Berlin   International
Workshop on `Text  Representation and  Domain Modelling'  (October 1991)
and of a  KIT Projekt  Kolloquium (TU  Berlin; February  1992) have also
helped  the  development   of  the   discussion  significantly.      The
particular  opinions  expressed  in  the  paper,  and  especially  their
deficiencies, remain however my responsibility.

\end{document}